\documentclass[twoside]{article}

\usepackage[accepted]{aistats2026}
\usepackage[round,authoryear]{natbib}

\usepackage[utf8]{inputenc} 
\usepackage[T1]{fontenc}    
\usepackage{hyperref}       
\usepackage{url}            
\usepackage{booktabs}       
\usepackage{amsfonts}       
\usepackage{nicefrac}       
\usepackage{microtype}      
\usepackage{xcolor}         
\usepackage{wrapfig}
\usepackage{multirow}

\usepackage{placeins}   
\usepackage{stfloats}   
\usepackage{flafter}    

\usepackage{mathtools}
\usepackage{comment}
\usepackage{algorithm2e}
\usepackage{amsthm}
\usepackage{graphicx}
\graphicspath{ {./figures/} }
\usepackage{amssymb}
\newtheorem{lemma}{Lemma}
\newtheorem{definition}{Definition}
\newtheorem{remark}{Remark}
\newtheorem{corollary}{Corollary}
\usepackage{afterpage}
\usepackage[capitalize]{cleveref}
\newcommand{\unc}[1]{\textnormal{\scriptsize$\pm$#1}}

\let\oldeqref\eqref 
\renewcommand{\eqref}[1]{Equation~\oldeqref{#1}} 
\makeatletter
\renewcommand{\citet}[1]{%
  \@ifundefined{r@cit@#1@author}
    {\citep{#1}}
    {\originalcitet{#1}}
}
\makeatother

\renewcommand{\arraystretch}{0.9}

\begin{document}

%

%

\twocolumn[

\aistatstitle{Unsupervised Ensemble Learning Through Deep Energy-based Models}

\aistatsauthor{ Ariel Maymon \And Yanir Buznah \And  Uri Shaham }

\aistatsaddress{ Bar-Ilan University } ]

\begin{abstract}
Unsupervised ensemble learning emerged to address the challenge of combining multiple learners' predictions without access to ground truth labels or additional data. This paradigm is crucial in scenarios where evaluating individual classifier performance or understanding their strengths is challenging due to limited information. We propose a novel deep energy-based method for constructing an accurate meta-learner using only the predictions of individual learners, potentially capable of capturing complex dependence structures between them. Our approach requires no labeled data, learner features, or problem-specific information, and has theoretical guarantees for when learners are conditionally independent. We demonstrate superior performance across diverse ensemble scenarios, including challenging mixture of experts settings. Our experiments span standard ensemble datasets and curated datasets designed to test how the model fuses expertise from multiple sources. These results highlight the potential of unsupervised ensemble learning to harness collective intelligence, especially in data-scarce or privacy-sensitive environments. 
\end{abstract}
\section{INTRODUCTION}\label{sec:introduction}
\includecomment{Ensembling is actually everywhere}
Ensemble learning manifests in various real-world applications, from crowd-sourced platforms to healthcare expert consultations and multi-sensor systems in autonomous vehicles. These diverse scenarios all seek to leverage shared and complementary information from multiple learners to create a more accurate and resilient meta-learner ~\citep{dong2020survey,ganaie2022survey}. The importance of ensemble methods has grown significantly with the rapid advancement of deep learning techniques, each offering unique strengths across various fields ~\citep{yangdeep2023survey}. 

Unsupervised ensemble learning addresses the critical challenge of combining multiple learners without access to ground truth labels, essential when labeled data is scarce~\citep{zhang2022knowledge}. This is particularly valuable when there is no clear metric to assess individual learner performance or adequacy. Within this domain, an important case arises when even additional data or learner features to learn from are unavailable. This constraint can be relevant in various scenarios, such as federated learning (due to privacy concerns)~\citep{mcmahan2017communicationefficient}, mixture-of-experts models, and multi-modal environments where different data types (e.g., images, text, video) make traditional representation fusion methods challenging to use~\citep{liang2024multimodal-survey}. Another rising multi-modal area is foundation models, where most of the work focuses on aggregating different modalities at the representation level. By focusing solely on the learners' predictions, such methods can offer a flexible framework for fusing knowledge across diverse models and domains, especially in such constrained settings. 

However, in these settings, existing unsupervised ensemble methods often leave potential performance untapped due to their limiting assumptions about learner relationships. Traditional approaches like majority voting are effective mostly when learners are independent and perform similarly~\citep{shaham2016deep}, while more sophisticated models such as the Dawid-Skene model~\citep{dawid1979maximum} impose restrictive conditional independence assumptions, which might perform sub-optimally when they do not hold. These constraints can fail to capture the intricate inter-connections that frequently exist among diverse learners in real-world applications, hindering the full exploitation of their collective knowledge. This unrealized potential motivates our approach to harness the power of deep learning to try to model complex learner interactions more effectively.

We argue that deep learning architectures, with their ability to disentangle latent representations and model complex interactions, provide a promising solution to this challenge. Motivated by this insight, we seek to develop a novel unsupervised ensemble learning method that reformulates the Dawid-Skene model as an energy-based model. This allows us to incorporate deep layers, creating a flexible and end-to-end trainable framework that captures sophisticated cross-learner relationships while building upon the theoretical foundations of the original model. 

Our contribution can be summarized as: (1) We present the iRBM, an identifiable variant of a Fully Multinomial RBM and show its equivalence to the Dawid-Skene model with theoretical convergence guarantees. (2) We use the iRBM to propose DEEM, a deep model that tackles the conditional independence assumption in an unsupervised way. (3) We effectively demonstrate our model's improved performance over existing baselines, in varying real-world ensemble scenarios. A PyTorch implementation will be provided upon acceptance to facilitate further research. 
\section{RELATED WORK}\label{sec:related-work}
\paragraph{Ensemble Using Additional Information} This line of work focuses on learning true labels by utilizing information additional to the learners predictions. For instance, \cite{chan2022synthetic} assume access to information about the learners' densities or the data on which they were trained. Similarly, \cite{rodrigues2018deep},\cite{chen2020structured}, \cite{chu2021conal} and \cite{cao2019maxmig} primarily rely on training data (the actual example) to construct their ensemble predictions. Our method differs from the above as it does not require additional inputs, focusing solely on learners' predictions to generate labels.

\paragraph{Ensemble Using Learners Predictions} Several other methods base their ensemble on extracting useful knowledge from the learners' different predictions alone. Possibly the first work to consider an unsupervised ensemble setup (in crowd-sourced applications) is the seminal Dawid-Skene (DS) model presented by \cite{dawid1979maximum}, which estimates the maximum likelihood of the latent true labels via an EM algorithm, assuming conditional independence between all learners given the label. Many subsequent methods have built atop of the DS model, to improve its efficiency or extend it. For instance, some exemplary works try to adapt this scheme to adversarial learners \citep{ma2020adversarial}, or identify the Dawid-Skene model using second-order statistics \citep{ibrahim2019crowdsourcing,ahsen2019}. The DS model is a valuable setup as it provides convergence guarantees to the true labels, but its full applicability is questionable since the conditional independence condition tends not to hold in practice \citep{jaffe2016unsupervised}.

Additional works considered relaxing the conditional independence assumption up to a certain degree. \cite{jaffe2016unsupervised} assume conditional independence might be present among some classifiers, given their dependence structure can still be described by a tree of depth two, while work by \cite{tenzer2022crowdsourcing} allows up to order of $\frac{2}{m}$ correlated error pairs in the covariance matrix for $m$ classifiers. The closest to ours is \citep{shaham2016deep}, that formulates the DS model as a binary Restricted Boltzmann Machine (RBM), and accommodates dependence among learners through stacked RBMs, trained layer by layer. \cite{li2019exploiting} model the dependence structure by a Markov network, assuming the correlations parameters are drawn from a Gaussian.

Unlike these methods, we propose a simple deep learning architecture in order to accommodate possible dependencies between learners, learning their relations in an unsupervised way. Although our guarantees only hold for the simple case, we empirically show our method outperforms other baselines when this assumption does not hold. Our method also trains end-to-end using an energy-based loss, is inherently multi-class, trains in mini-batches and outputs a classifier that can be used for new unseen data. 

\paragraph{Relation to Programmatic Weak Supervision} This framework \citep{ratner2016dataprogramming} uses multiple weak labeling functions (LFs) on unlabeled data to recover true labels for training a downstream model~\citep{zhang2022survey-pws,wu2023hlm,tonolini2023robust}. PWS differs from our approach in key aspects: PWS labeling functions may cover only subsets of data, while we assume all learners provide predictions for the entire dataset. Additionally, we employ an ensemble for immediate and future inference without further training, unlike PWS which aims to train a separate end model. Despite these differences, PWS represents the closest existing paradigm to our work, therefore we will empirically compare our method against relevant PWS techniques to evaluate its performance.

\section{BACKGROUND}
\label{section:background}
\subsection{Unsupervised Ensemble Learning}
We consider a multi-class classification setting. Let $X \in \{1,\dots,K\}^d, Y \in \{1,\dots,K\}$ be random variables. In this paper, the vector $X = (X_1, \ldots, X_d)^\top$ represents the predictions of $d$ learners or annotators on an instance, where $Y$ is the true label and $K$ is the number of classes.  The pair $(X, Y)$ has a joint distribution which is given by $p_\theta(X, Y) = p_\theta(Y)p_\theta(X|Y)$. The joint distribution $p_\theta(X, Y)$ is not known to us, neither are the marginals $p_\theta(X)$, $p_\theta(Y)$.

Let $(x^{(1)}, y^{(1)}), \ldots,(x^{(n)}, y^{(n)})$ be $n$ i.i.d. samples from $p_\theta(X, Y)$. In unsupervised ensemble learning, we observe $\{x^{(i)}\}_{i=1}^n$ and aim to recover $\{y^{(i)}\}_{i=1}^n$. We use the words classifiers or learners interchangeably and the notation $p(x)$ as a shorthand for $\Pr(X=x)$, the probability of the variable $X$ to be the value $x$.

\subsection{The Conditional Independence Model}
\cite{dawid1979maximum} assumed that the conditional distribution $p_\theta(X|Y)$ of the model factorizes, i.e.,
\begin{equation}
p_\theta(X|Y) = \prod_{i=1}^d p_\theta(X_i|Y). \label{eq:ci-model}
\end{equation}
\eqref{eq:ci-model}, also known as the conditional independence model, is fully parameterized by $\theta =
(\{\psi_{ilm} : i = (1, \ldots, d), l = (2,\dots,K), m = (1,\dots,K)\}, \{\pi_t : t = 2, \ldots, K\})$,
where
\begin{align*}
\psi_{ilm} &= \Pr(X_i = l|Y = m),
&\pi_t &= \Pr(Y = t).
\end{align*}
\begin{remark}\label{ci-model-remark}
By definition, $\psi_{i1m} = 1 - \sum_{l\neq 1} \psi_{ilm}$ for every pair of possible $(i,m)$ indices. Likewise, $\pi_1 = 1 - \sum_{t\neq 1} \pi_t$. 
\end{remark}
\subsection{Fully Multinomial RBM}
\begin{figure}[t]
    \centering
    \includegraphics[width=1.\linewidth]{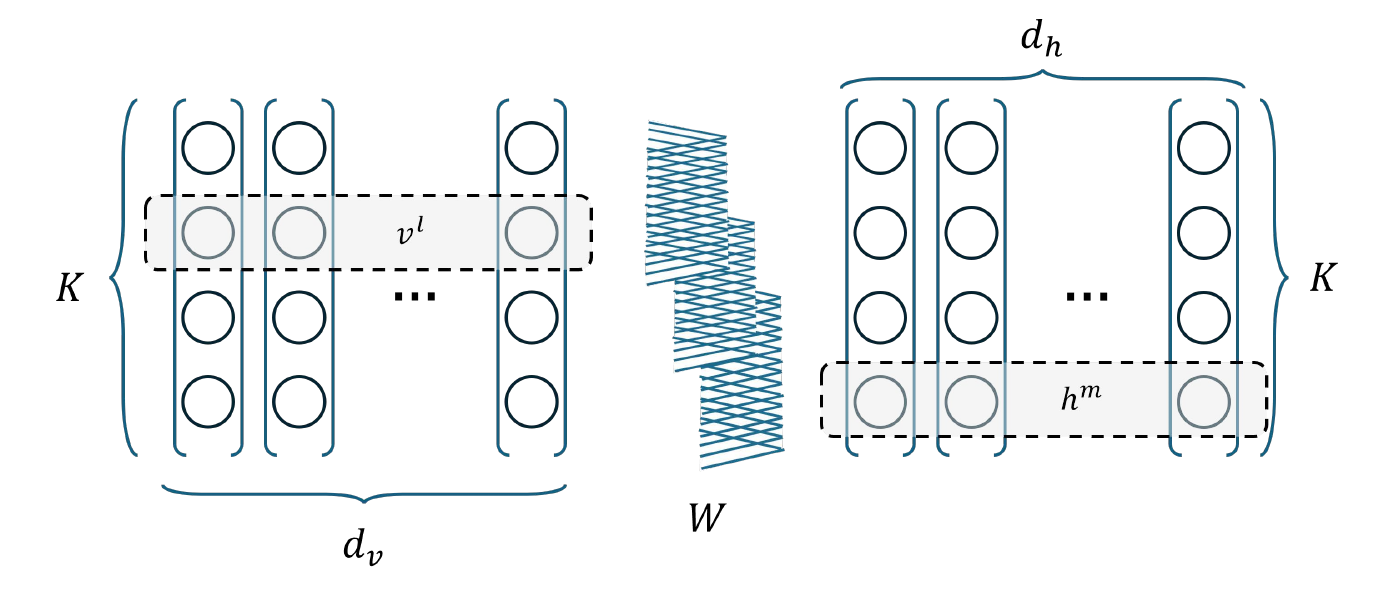}
    \caption{Fully-Multinomial RBM, with $d_v,d_h$ multinomial units, each with size $K$. The weights tensor $W$ connects from every node in one layer to all nodes in the other, and vice versa.}
    \label{fig:fm-rbm}
\end{figure}
A Fully Multinomial Restricted Boltzmann Machine (FM-RBM), depicted in Figure \ref{fig:fm-rbm}, extends the binary RBM to accommodate multinomial inputs and outputs. This bipartite graph generative Energy-Based Model (EBM) consists of a visible layer $V$ with $d_v$ units and a hidden layer $H$ with $d_h$ units. Each unit is represented as a one-hot encoded vector of size $K$, allowing for $K$ possible discrete states\footnote{While a general Fully Multinomial RBM can have different numbers of multinomial states for visible and hidden units, we use $K$ for both, as for our task it is sufficient. This simplification aids understanding without loss of generality.}. For information about RBMs and energy-based models, please see \autoref{background-appendix}.

The FM-RBM is parameterized by $\lambda = (W,a,b)$, where $W$ is a 4-way tensor of sizes $(K,K,d_v,d_h)$ representing the weights, while $a$ and $b$ are matrices of the visible and hidden biases, with sizes $(K,d_v)$ and $(K,d_h)$, respectively.  
For clarity, we use subscripts $i,j$ for the visible and hidden units indices, and superscripts $l,m$ to denote the classes of the visible and hidden multinomial units throughout this paper.

Each FM-RBM configuration $(V = v, H = h)$ is associated with its energy: 
\begin{align}
\label{eq:fm-rbm-energy}
   &E_\lambda(v,h) = \left(\sum_{l} {a^l}^T v^l + \sum_{m} {b^m}^T h^m  + \sum_{l,m} {v^l}^T W^{l,m} {h^m}\right)
\end{align}
which defines its probability as:
\[
p_\lambda(V=v,H=h) = \frac{e^{-E_\lambda(v,h)}}{Z(\lambda)}
\]
With $Z = \sum_{v,h} e^{-E_\lambda(v,h)}$ as the partition function. Note that $v$ and $h$ are matrices (of sizes $(K,d_v)$, $(K,d_h)$). $v^l,h^m$ are vectors that represent the $l$-th class in the visible layer and the $m$-th class in the hidden layer.

\eqref{eq:fm-rbm-energy} implies the conditional probabilities of the visible and hidden units can be represented as softmax functions (\autoref{fm-rbm-appendix}):  
\begin{align*}
    p(v_i=\mathbf{e}_k|h) &= \frac{e^{a^{k}_i + \sum_{m}\sum_{j} w^{km}_{ij} h^m_{j}}}{\sum_{l=1}^{K} e^{a^{l}_i + \sum_{m}\sum_{j} w^{lm}_{ij} h^m_{j}}}\\
    p(h_j=\mathbf{e}_k|v) &= \frac{e^{b^{k}_j+\sum_{l}\sum_{i} w^{lk}_{ij} v^l_{i}}}{\sum_{m=1}^{K} e^{b^{m}_j+\sum_{l}\sum_{i} w^{lm}_{ij} v^l_{i}}},
\end{align*}
where $\mathbf{e_k}$ is a one-hot vector, with 1 at its $k$-th dimension and 0 at the rest. $\mathbf{e_k} \in (\mathbf{e_1},\dots,\mathbf{e_K})$. 

Given $\{x_i\}_{i=1}^n \sim p_\theta(X)$ i.i.d. training data, the FM-RBM parameters are tuned to maximize the log-likelihood of the training data, using energy-based optimizations. In our case, each $x_i$ is encoded as a 2D matrix of dimension $(K,d_v)$, that contains one-hot representations of all labels in the original predictions vector (See \autoref{fm-rbm-example-appendix} for an example).

\section{PROPOSED APPROACH}
\label{section:proposed-approach}
This section presents a comprehensive development of our proposed method. We begin in \autoref{subsec:irbm} by addressing an identifiability issue of the FM-RBM and introducing the identifiable FM-RBM (iRBM) model. We then establish the equivalence between the iRBM and the Dawid-Skene (DS) model, clarifying their relationship. In \autoref{subsec:unsupervised_ensemble}, we show guarantees for the recovery of the true posterior by the iRBM under the conditional independence assumption. Finally, \autoref{subsec:deem} introduces our full model, where we introduce a deeper structure as a heuristic for cases when the conditional independence assumption is not satisfied, thereby enhancing the model's capacity to capture complex inter-connections between learners.

\subsection{The \lowercase{i}RBM as a DS Model}
\label{subsec:irbm}
The softmax function is translation invariant, meaning that adding a constant to all arguments yields the same probability distribution. Since we use it for the FM-RBM multinomial units, the model is not identifiable as its parameters cannot be uniquely determined. This issue prevents us from constructing a map between the unique solutions of the FM-RBM and the conditional independence model, necessitating a modification to the FM-RBM structure, where every multinomial unit has one fixed coefficient.

We address this by defining the Fully Multinomial Identifiable RBM (iRBM), which we will use to prove the equivalence result with:
\begin{definition}{FM-iRBM}\label{fm-irbm}
    A Fully-Multinomial Identifiable RBM (iRBM) is defined as a Fully-Multinomial RBM parameterized by $\lambda = (W,a,b)$, with the same energy and probability distribution of the FM-RBM as in  \eqref{eq:fm-rbm-energy}. In addition:
$\begin{cases}
a_i^l = 0,\ w_{ij}^{lm} = 0,\ b_j^m = 0 & \text{if } l = 1 \text{ xor } m = 1 \\
w_{ij}^{lm} = 1                         & \text{if } l = 1 \text{ and } m = 1 \\
\end{cases}$\\
are constants\footnote{The fixed parameters can be any scalar, but we chose these values for simplicity and consistency with our initialization scheme.}. Refer to \autoref{irbm-visualization} for illustration.
\end{definition}
To show the equivalence of the iRBM to the conditional independence model, we proceed to present a bijective map from the parameters $\lambda$ of an iRBM with a single hidden unit, to the parameters $\theta$ of the conditional independence model, so that the joint distribution specified by the iRBM is equivalent to that of the conditional independence model.
\begin{lemma}
 \label{lem:ci-rbm}
The joint probability $p_\lambda(V=x,H=y)$ of an iRBM with the parameters $\lambda = (W,a,b)$ with $d_v=d$ and $d_h=1$, is equivalent to the joint probability $p_\theta(X=x,Y=y)$ of the conditional independence model with the parameters $\theta = (\{\psi_{ilm}\},\{\pi_t\})$ given by:
 \begin{align*}
    &\psi_{ilm} \equiv \sigma(\mathbf{z})_{ilm} ,
    &\pi_t \equiv \frac{\sum_v e^{b^t + \sum_{l} v^l W^{l,t}}}
{\sum_{v,h\in \{\mathbf{e}_1,\dots,\mathbf{e}_K\}} e^{b^m + \sum_{l} v^l W^{l,m}}}
 \end{align*}
where $\sigma$ is the softmax function, $z_{ilm} = (a_i^l + w_i^{lm})$, and $\mathbf{z} = (z_{i1m},\dots,z_{iKm}) \in \mathbb{R}^K$. The map $\lambda \longmapsto \theta$ is a bijection.
\end{lemma}
Proof can be found in \autoref{ci-proof-appendix}.

It can be seen that the effective number of free parameters of the iRBM and the conditional independence model are both exactly equal to $(dK+1)(K-1)$ parameters.

Note that the result proven by \cite{shaham2016deep} for the binary RBM can be considered a special case of our Lemma \autoref{lem:ci-rbm} where $K=2$ and $z_{ilm} = -(a_i^l + w_i^{lm})$ for $l=1$ or $m=1$ and $b_1 = -b_1$, which will zero out the logits of all exponents of the first hidden and visible units, essentially yielding the sigmoid function as the conditional probability of the RBM for 2 classes.
\begin{remark}
\label{remark:iden}
Let $\{x_i\}_{i=1}^n$ $be$ observed data from the conditional independence model, specified by $p_{\theta}$. Given $\theta$, and under the mild assumptions that $\forall j \in \{1,\dots,d\},  X_j \not\!\perp\!\!\!\perp Y$ (every classifier is not independent of Y. i.e., not merely a random guess), and $d \geq 3$, then the conditional independence model is a special case of the tree model presented by \cite{chang1996full}, which was proved identifiable under these conditions. Therefore, the conditional independence model is also identifiable.
\end{remark}
\begin{corollary}
\label{corollary:fm-irbm-identifiable}
Based of Lemma \autoref{lem:ci-rbm}, and Remark \autoref{remark:iden}, it follows that the iRBM is identifiable as well.
\end{corollary}
\subsection{\lowercase{i}RBM for Ensemble Learning}
\label{subsec:unsupervised_ensemble}
We now show how the posterior probability $p_\theta(Y=y|X=x)$ can be indeed recovered by a iRBM with a single hidden unit in a consistent manner, given some $\{x_i\}_{i=1}^n \sim p_\theta(X)$ observed data from the conditional independence model defined in \eqref{eq:ci-model},  summarizing the main result of the equivalence. We prove this corollary similarly to \citep{shaham2016deep}, and present the proof in \autoref{mle-proof-appendix} for completeness.
\begin{corollary}\label{corollary:posterior}
For any $x,y$, the iRBM posterior probability $p_{\hat{\lambda}_\text{MLE}}(H=y|V=x)$ converges to the true posterior $p_\theta(Y=y|X=x)$, as $n\to\infty.$
\end{corollary}
Due to the energy-based nature of the iRBM and its architecture, two notable considerations arise regarding the limitations of implementing this equivalence in practice:
\begin{remark}
\label{remark:mapping}
The parameters’ identifiability is up to a global permutation of the classes; Meaning that one might recover a map, where the multinomial output units of the iRBM correspond to a permutation of the conditional independence model class labels, potentially relabeling the classes. Formally, there is a bijective mapping $\phi : \{1, ..., K\} \longrightarrow \{1, ..., K\}$  such that $p_\theta(Y = y|X=x) = p_\lambda(Y = \phi(y)|V=x)$.
\end{remark}As on average, the $X_j$s are more accurate than a random guess, this can be resolved by solving the assignment problem between the predictions and the majority vote decision. As will be discussed later, since we initialize the iRBM with a majority-vote (MV) initialization, the permutation often defaults to the identity mapping in our implementation.
\begin{remark}
\label{remark:achieving-mle}
Corollary \autoref{corollary:posterior} is based on the assumption that we found the iRBM $\hat{\lambda}_\text{MLE}$. In practice, achieving the MLE is difficult due to the intractability of the EBM true gradient, resulting in maximizing a proxy to the gradient. More so, there is no guarantee that after training an iRBM one can obtain the MLE, since its likelihood function is not concave.
\end{remark}
\subsection{Deep Energy Ensemble Models}
\label{subsec:deem}
\begin{figure}[t]
    \centering
    \includegraphics[width=1.\linewidth]{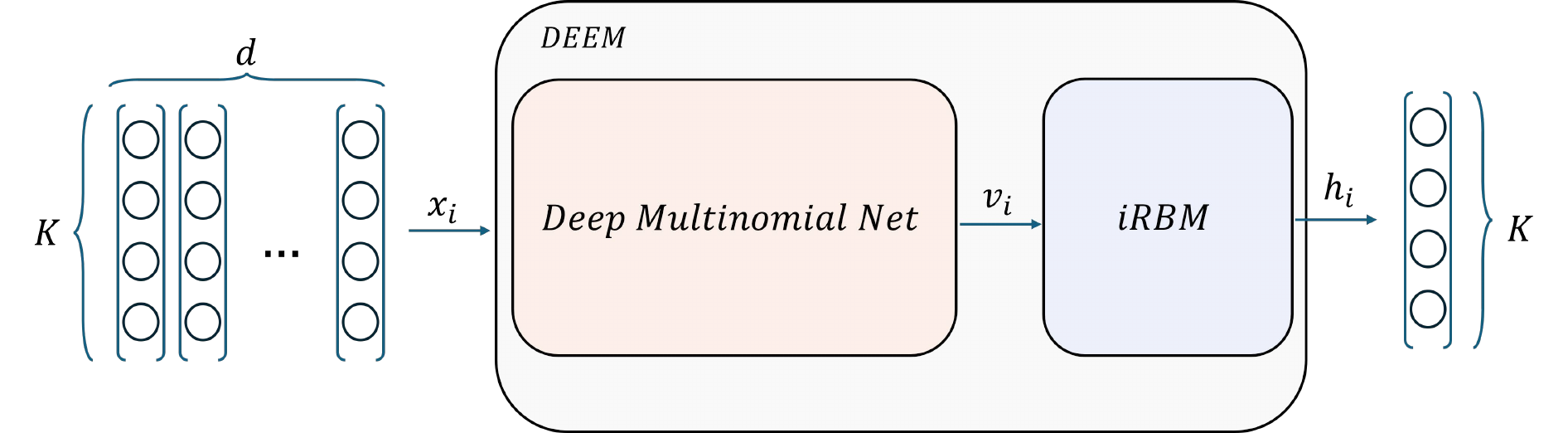}
    \caption{The DEEM model. The sample $x_i$ goes through the deep network, to acquire the visible input $v_i$ to the iRBM layer, which then forwards it to get $h_i$. In training, $v_i,h_i$ are used to compute the gradient which then propagates backwards. When inferencing on $x_i$, the prediction $h_i$ is mapped to the correct class and returned.}
    \label{fig:deem}
\end{figure}
After establishing the equivalence between the iRBM and the conditional independence model, we address the limitation of the conditional independence between input learners, which is often violated in practice and might yield sub-optimal results for the iRBM predictions.

To overcome this, we propose a Deep Energy-based unsupervised Ensemble Model (DEEM), depicted in \autoref{fig:deem}, extending iRBM with deep layers. DEEM propagates input through Multinomial layers, with an iRBM predictor estimating $p_\theta(Y|X)$ via Corollary \autoref{corollary:posterior}.

The incorporation of deep layers and non-linear activation functions is motivated by their ability to form disentanglement of latent variations in the inputs \citep{mehta2014exact,tishby2015deep,Shwartz2017Opening} and their richer expressive power \citep{montúfar2014number,eldan2016power,vardi2022width}. Intuitively, we can view the propagation through the layers as a process that preserves latent label information while reducing conditional dependence between the input features, enabling DEEM to better approximate complex posteriors $p(Y|X)$ with non-trivial learner relations.

Post-training, we use majority vote decision for class-mapping (Remark \autoref{remark:mapping}), enabling prediction on seen and unseen examples. Key implementation details follow, with additional information in \autoref{additional-training-info-appendix}.
\paragraph{The Multinomial Layer}
The Multinomial layer is a deep network component we design to keep the probabilistic nature of DEEM and our input, while allowing easy forward propagation and the benefits of the deep structure as discussed earlier. It can be seen as an extension to the regular linear layer but with multinomial inputs and outputs.

Formally, a Multinomial layer consists of a weight 4-way tensor $W$ of size $(K,K,d,d)$ and bias $b$ of size $(K,d)$ , Where $K$ is the multinomial unit size, and $d$ is the input and output sizes\footnote{In principle, this layer can take any $d_\text{in},d_\text{out}$ sizes, but we found that assigning them equal results in more robust training of the model.}.

Similarly to the FM-RBM, An example $x$ sized $(K,d)$, is forward propagated through the layer as follows:
\begin{align*}
z^m_{j} &=\sum_{l=1}^{K} \sum_{i=1}^{d} w^{lm}_{ij} x^l_{i} + b^m_{j} ,
&a_{j} &= f(z^1_{j},\dots,z^m_{j})
\end{align*}
to give $z$ of sizes $(K,d)$, where $f$ is an activation function, $i,j$ are the indices of the input and output dimensions and $l,m$ are the indices for the multinomial dimensions. Conceptually, each layer propagates a set of one-hot\footnote{Due to its structure, this architecture also supports soft-label injection naturally. We focus one-hot representations for clarity.} predictions to new latent set of predictions.

We employ Sparsemax \citep{martins2016sparsemax} as the activation function for the multinomial layer. This 'sparser' alternative to softmax assigns probabilities only to sufficiently large logits, zeroing the rest. This approach yielded better results than softmax, likely due to its handling of our one-hot input data.
\subsubsection{Training and Inference Details of DEEM}
DEEM is trained in a standard end-to-end fashion, calculating the classic energy loss $\log p_w(x) = -E_w(x) - \log Z(w)$ based on \eqref{eq:grad-term-ebm}, employing the Deep Langevin Proposal sampler\citep{zhang2022langevinlike}. Post-training, the Hungarian algorithm \citep{kuhn1955hungarian} compares DEEM predictions with the majority vote decision to obtain a class map for inference. \autoref{algorithm:deem} provides a detailed pseudo-code of the training and inference procedures. 

Given a trained DEEM and a sample $x \sim p_\theta(X)$, its label $y$ is estimated by propagating $x$ through the whole model. The output is then decided by taking the $\arg \max p_\lambda(h|x)$, and mapping using the class map function. Note that due to the stochastic nature of DEEM, one can also estimate the label by sampling from $p_\lambda(h|x)$, propagating $x$ through the network several times and averaging the outputs to obtain $\mathbb{E}(Y|X=x)$. Experimentally, we found the $\arg\max$ estimate to be sufficiently effective in most cases.
\begin{algorithm}
\setlength{\textfloatsep}{6pt plus 2pt minus 2pt}  
\setlength{\intextsep}{6pt plus 2pt minus 2pt}
\setlength{\abovecaptionskip}{4pt}
\setlength{\belowcaptionskip}{2pt}
\caption{DEEM: Energy-Based Model for Classification}
\label{algorithm:deem}
\SetAlgoLined
\smaller
\DontPrintSemicolon
\KwIn{$\mathbf{x}$: batch of one-hot encoded inputs, $\lambda$: model parameters}
\KwOut{$\phi$: class mapping function}
\SetKwProg{Proc}{Procedure}{}{}
\Proc{Train}{}{
Initialize $\lambda$ with appropriate priors\;
\While{convergence criterion not met}{
Sample negative examples $\mathbf{x}_{\text{neg}}$\;
Propagate $\mathbf{x}$ and $\mathbf{x}_{\text{neg}}$ through the model to obtain $(\mathbf{v},\mathbf{h})$ and $(\mathbf{v}_{\text{neg}},\mathbf{h}_{\text{neg}})$\;
$E_{\text{pos}} \gets \text{Energy}(\mathbf{v},\mathbf{h}, \lambda)$,  $E_{\text{neg}} \gets \text{Energy}(\mathbf{v}_{\text{neg}},\mathbf{h}_{\text{neg}}, \lambda)$\;
$\mathcal{L} \gets E_{\text{pos}} - E_{\text{neg}}$\;
Freeze iRBM constants\;
Update $\lambda$ using gradient of $\mathcal{L}$\;
}
$\hat{Y} \gets \text{Model}(\mathbf{x}, \lambda)$, $Y_{\text{MV}} \gets \text{MajorityVote}(\mathbf{x})$\;
$\phi \gets \text{HungarianAlgorithm}(\hat{Y}, Y_{\text{MV}})$\;
\Return $\lambda,\phi$
}
\BlankLine
\SetKwProg{Proc}{Procedure}{}{}
\Proc{Inference}{}{
\KwIn{$\mathbf{x}$: input sample, $\lambda$: trained model parameters, $\phi$: class mapping function}
\KwOut{$y$: predicted class label}
$\mathbf{h} \gets \text{Propagate}(\mathbf{x}, \lambda)$\;
$y \gets \phi(\arg\max_k(\mathbf{h}_k))$\;
\Return $y$
}
\end{algorithm}
\section{EXPERIMENTAL RESULTS}
\label{section:experimentalResults} 
In this section, we present our method performance. First, we show that the iRBM accurately reconstructs the parameters of the original conditional independence model, and that the multinomial layers of DEEM empirically promote conditional independence. Then, we benchmark DEEM against other baselines on several simulated and real-world datasets. Lastly, we evaluate DEEM on unique ensemble scenarios, deriving insights for its use in practice. Detailed information about the datasets in this section can be found in \autoref{datasets-appendix}.
\subsection{Recovery of the DS Parameters}
 \begin{figure}
     \centering
     \includegraphics[width=1.\linewidth]{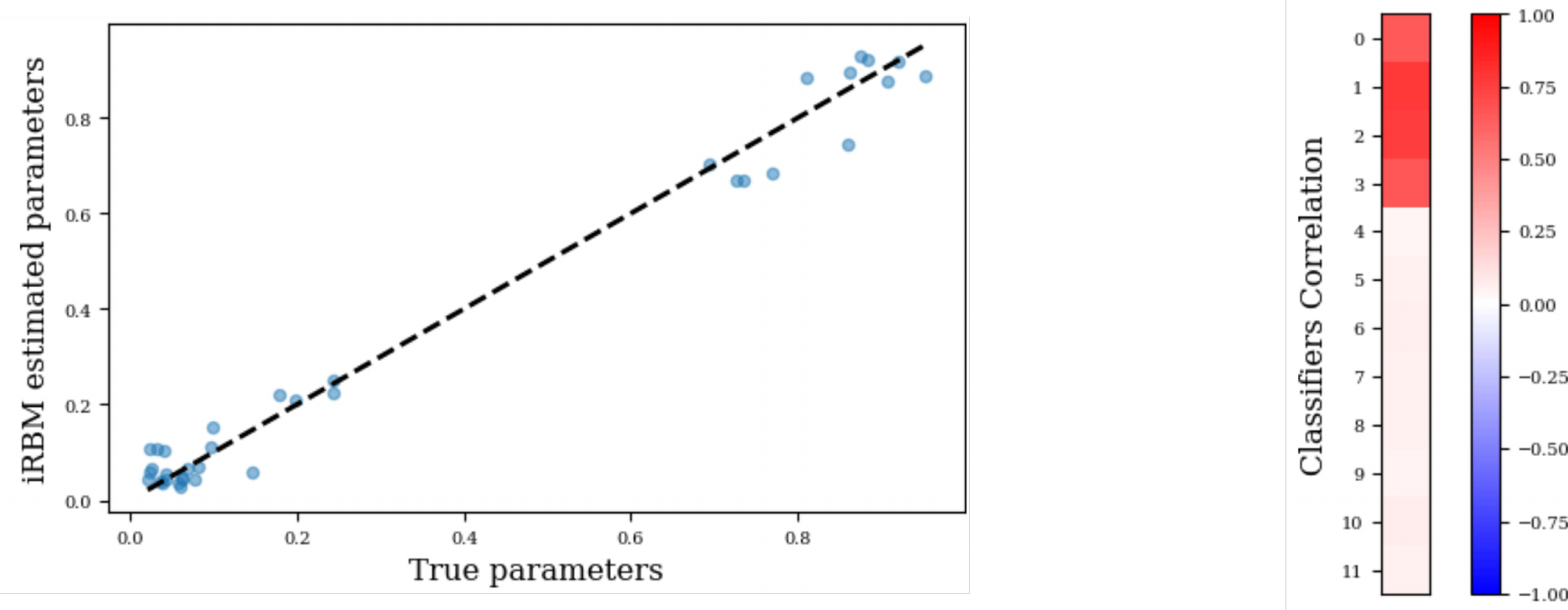}
     \caption{Recovery graph (left) and a weight correlation heatmap (right). In the recovery graph, each circle is the parameter value from the DS model $(\psi_{ilm} , \pi_t)$ (X-axis), and its corresponding iRBM parameter value (Y-axis), using the map outlined in Lemma~\autoref{lem:ci-rbm}. The closer to the dotted identity line, the better the recovery (as it means they have the same value). It can be seen that the iRBM recovers the DS model original parameters correctly. The heatmap shows each classifier's correlation with the iRBM final prediction, which shows the iRBM was able to tell the classifiers that benefited the prediction, and rule out the rest (which were random guesses).}
     \label{fig:ci-irbm}
 \end{figure}
To test for our model's capability to correctly recover the true DS model parameters, we trained an iRBM on a simulated dataset (CondInd), which consists of $K=3$ classes and conditionally independent classifiers, where the first four classifiers predictions are drawn from the original labels with a small perturbation, and the rest are all random guesses. 

We show the results in Figure \ref{fig:ci-irbm}, where we can observe how the iRBM successfully recovers the original parameters that created the dataset. Additionally, we can see the correlation with the iRBM weight vector to the prediction, suggesting it was able to distinguish 'true' classifiers from those that do not benefit the prediction.
\subsection{Disentanglement Through the Multinomial Layers}\label{disentaglement-multinomial}
Upon the addition of the multinomial layers to mitigate conditional dependencies between learners, we empirically validate the claim by the analysis below, conducted on the trained DEEM of the MnistE dataset: 

At each layer of DEEM, and for each true label (used only for analysis), we compute the mutual information (MI)\footnote{We computed the MI using mutual\_info\_score from scikit-learn, which performs binning via empirical co-occurrence counts.} between classifier outputs as a proxy for conditional independence. The input to the MI computation is the argmax prediction of each classifier (feature), treated as a discrete random variable. 
\begin{table}[h]
\centering
\caption{Max MI and Frobenius Norm mean and std per true class, across the multinomial layers.}
\label{tab:maxmi_frob}
\resizebox{\linewidth}{!}{
\begin{tabular}{lcc}
\toprule
\textbf{Layer} & \textbf{Max MI} & \textbf{Frobenius Norm} \\
\midrule
Layer 0 (Input)      & $0.827 \pm 0.294$ & $4.428 \pm 1.796$ \\
Layer 1              & $0.981 \pm 0.272$ & $3.504 \pm 1.439$ \\
Layer 2 (iRBM input) & $0.314 \pm 0.068$ & $2.687 \pm 0.985$ \\
\bottomrule
\end{tabular}}
\end{table}
For each true class and each layer, we construct a full $d \times d$ MI matrix which we summarize below using the maximum MI and Frobenius norm (Excluding diagonal entries).

\begin{figure}[!ht]
    \centering
    \includegraphics[width=1\linewidth]{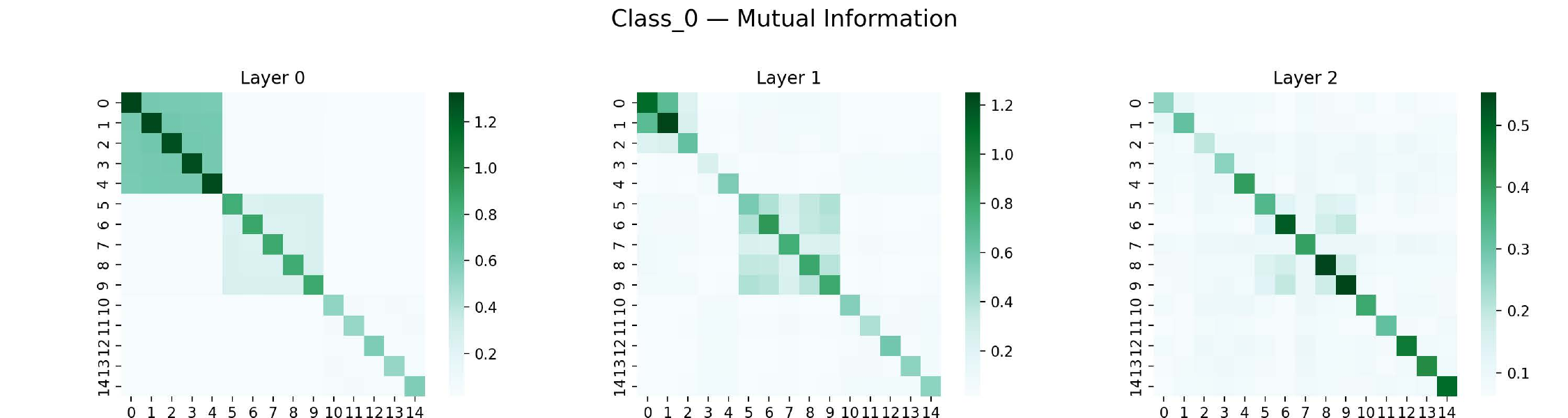}
    \caption{The conditional mutual information matrices of a trained DEEM with 2 multinomial layers. Starting from the input layer in the leftmost column, it can be seen how the mutual information is gradually reduced as we progress through the multinomial network, until we get disentangled features as input for the iRBM component. This is a typical plot from one particular true label class subset of the data.}
    \label{fig:maxmi_frob_paper}
\end{figure}

\begin{table*}[!hb]
    \centering
    \small
    \caption{Accuracy Performance. Table shows methods performance on each of the datasets. BestClf and AvgClf is presented as a reference. Methods without standard deviations are either deterministic by design or implementation (i.e., EBCC where one run is selected based on highest ELBO). Terms $det$ and $sto$ are the deterministic and stochastic variants of DNN, respectively.}
    \label{fig:main-results-ni}
    \resizebox{\textwidth}{!}{%
    \setlength{\tabcolsep}{4pt} 
    \renewcommand{\arraystretch}{0.9} 
    \begin{tabular}{lcccccccc}
    \toprule
        & Tree3k & MnistE & PetFinder & CSGO & MicroAgg2 & EyeMovem & ArtiChars & GesturePhsm \\ 
    \midrule
    BestClf & 87.62 & 86.79 & 94.09 & 90.13 & 63.39 & 74.75 & 92.87 & 69.98 \\
    AvgClf & 73.59\unc{10.47}& 65.72\unc{16.0}& 65.68\unc{17.5}& 78.79\unc{8.31}& 60.3\unc{2.49}& 63.5\unc{9.52}& 70.39\unc{16.36}& 60.69\unc{6.46}\\ \midrule[0.9pt]
    MV & 94.63 & 82.41 & 77.02 & 85.34 & 61.85 & 71.32 & 79.04 & 64.23 \\ 
    L-SML & \textbf{95.67} & 85.63 & 79.57 & \textbf{88.82} & 62.80 & 72.45 & 79.69 & 65.38 \\ 
    DS & 95.29 & \underline{92.63} & \textbf{80.15} & 86.94 & \underline{63.01} & \underline{73.44} & 82.30& 64.72 \\ 
    EBCC & 94.98 & 87.92 & 78.58 & 86.79 & 62.82 & 73.26 & \textbf{84.10} & 63.43 \\ 
    HLM & 94.65 & 85.21 & 79.03 & 87.08 & 62.93 & 71.09 & 79.06 & 66.25 \\
    LA$_{2pass}$ & \underline{95.60}\,\unc{0.00} & 86.71\,\unc{0.05} & 78.05\,\unc{0.00} & 86.79\,\unc{0.00} & 61.95\,\unc{0.00} & 72.49\,\unc{0.01} & 79.95\,\unc{0.01} & 64.48\,\unc{0.00} \\ 
    FSquid & 94.35 & 77.35 & 69.50 & 82.58 & 61.10 & 72.18 & 78.91 & 62.03 \\ 
    DNN$_{sto}$ & 93.28\,\unc{1.73} & 77.73\,\unc{4.48} & 78.36\,\unc{0.27} & 82.79\,\unc{7.03} & 61.90\,\unc{0.88} & 68.48\,\unc{10.23} & \underline{83.53}\,\unc{1.37}& \underline{66.57}\,\unc{0.59}\\
    DNN$_{det}$ & 93.10\,\unc{1.73} & 78.63\,\unc{4.49} & 76.79\,\unc{0.93} & 82.56\,\unc{6.86} & 62.35\,\unc{0.21} & 67.44\,\unc{10.15} & 82.84\,\unc{2.30} & 65.88\,\unc{0.79} \\
    \midrule[0.5pt]
    iRBM & 95.23\,\unc{0.05} & 87.13\,\unc{0.04} & 77.97\,\unc{0.07} & 85.75\,\unc{0.14} & 61.86\,\unc{0.00} & 73.03\,\unc{0.10} & 79.22\,\unc{0.04} & 66.55\,\unc{0.05}\\
    \textbf{DEEM} & 95.52\,\unc{0.08} & \textbf{94.95}\,\unc{0.00} & \underline{79.84}\,\unc{0.08} & \underline{88.16}\,\unc{0.46} & \textbf{63.06}\,\unc{0.01} & \textbf{73.73}\,\unc{0.29} & 82.21\,\unc{0.33} & \textbf{67.00}\,\unc{0.51} \\ 
    \bottomrule
    \end{tabular}
    }
\end{table*}
These statistics reflect an approximation for the level of dependence among classifiers at each layer. As shown in Table~\ref{tab:maxmi_frob} and Figure~\ref{fig:maxmi_frob_paper}, both max MI and Frobenius norm decrease across layers, supporting our hypothesis that the learned representations are shifted towards conditional independence as we move through the multinomial network.
\subsection{Unsupervised Ensembles}
We evaluate our method on a carefully curated set of simulated and real-world datasets, designed to showcase its ability to capture complex learner interactions—a crucial aspect of high-quality ensembles. Our benchmark spans across diverse scenarios, including multi-modal cases. This enables a more rigorous assessment of the effectiveness of unsupervised ensemble methods across varied applications. 

\paragraph{Datasets} The datasets include Tree3K, a simulated dataset designed to test performance in presence of hierarchical conditional dependencies between learners. MnistE, constructed from three types of classifiers trained on MNIST data, emulates a standard expert ensemble scenario. For multi-modal evaluation, we incorporate the CS:GO dataset from MuG \citep{lu2023mug}, which combines text, image, and tabular features. This dataset's ensemble consists of uni-modal models trained using AutoGluon \citep{erickson2020autogluon}. Additionally, we include several tabular datasets derived from UCI \citep{dua2019uci}, utilizing predictions from diverse pre-trained tabular model architectures as provided by TabRepo \citep{salinas2023tabrepo}.

This carefully assembled collection presents a wide spectrum of ensemble configurations and classifier sets, offering a robust evaluation framework for unsupervised ensemble methods. Our datasets are particularly valuable for assessing performance where the conditional independence assumption is violated, a common limitation in real-world applications. Additional specifications and preprocessing steps are in \autoref{datasets-appendix}. 

\paragraph{Benchmarks} We compare our performance to classic methods such as majority-vote (MV) and DS\footnote{Using an enhanced implementation by~\cite{sinha2018fast}}~\citep{dawid1979maximum}, unsupervised ensemble models like L-SML~\citep{jaffe2016unsupervised}, DNN~\citep{shaham2016deep} and recent PWS and crowd-sourcing methods such as EBCC~\citep{li2019exploiting}, LA-pass~\citep{yang2024lapass},  HLM~\citep{wu2023hlm} and FlyingSquid~\citep{fu2020flyingsquid}, adapting to one-vs-all when necessary.
\begin{table*}[!t]
    \centering
    \small
    \caption{Accuracy on expert subsets (classes with oracle learners) and remaining data of each of the MoE datasets.}
    \label{table:moe}
    \resizebox{0.85\textwidth}{!}{%
    \small
    \setlength{\tabcolsep}{4pt} 
    \renewcommand{\arraystretch}{0.9} 
    \begin{tabular}{lcccccc}
    \toprule
    & \multicolumn{2}{c}{MnistE-4/7} & \multicolumn{2}{c}{MnistE-568} & \multicolumn{2}{c}{AmpData} \\
    Method & Expert& Remaining& Expert& Remaining& Expert& Remaining\\
    \midrule
    MV    & 84.82 & 82.98& 91.66 & 83.02 & 54.82 & 99.63\\
 L-SML& 88.32 & 86.20& 95.83& 87.55& 46.34&\textbf{99.92}\\
    DS    & 91.70 & \underline{92.52}& 95.07 & \underline{93.62} & \underline{89.29} & 97.64 \\
    LA$_{2pass}$   & 88.38\unc{0.0}& 86.73\unc{0.0}& 95.37\,\unc{0.0} & 89.57\,\unc{0.01} & 28.15\unc{0.05}& \underline{99.91}\unc{0.11}\\
    EBCC  & \underline{94.53}& 88.57& \underline{95.74} & 90.77 & 34.06 & 98.70 \\
    HLM   & 87.11 & 85.28& 92.59 & 87.19 & 60.28 & 98.36 \\
    \midrule
 iRBM & 92.28\unc{0.39}& 86.13\unc{0.25}& 96.07 \unc{0.10}& 88.89\unc{0.09}& 29.99\unc{0.98} &\textbf{99.92}\unc{0.01}\\
    \textbf{DEEM} & \textbf{95.27}\unc{0.08}& \textbf{94.69}\unc{0.0}& \textbf{96.07}\,\unc{0.42} & \textbf{94.29}\,\unc{0.76} & \textbf{96.63}\,\unc{0.44} & 96.17\,\unc{0.81} \\
    \bottomrule
    \end{tabular}
    }
\end{table*}
Additionally, we include the iRBM performance (to compare the benefits of DEEM), BestClf, representing the best individual classifier in the ensemble and AvgClf, the average individual classifier accuracy. While this information is not available in unsupervised settings, it provides readers with insight into the original strength of the ensemble's components, offering context for interpreting the unsupervised methods' performance. 

Table \ref{fig:main-results-ni} presents the performance on all datasets. The performance is averaged over five runs across all datasets, wherever applicable. DEEM has the highest average accuracy, being 0.6\% better on average than the second-best method, and is either the best or second-best performer on most of the benchmarks, as seen in the table.

\subsection{DEEM as a Mixture-of-Experts Model}
To test DEEM’s ability to uncover hidden specializations and expertise, we build three datasets:

\textbf{MnistE-4/7}, where two learners become an oracle on one of the digits, one being an expert on 4 and one on 7, additional to their performance. \textbf{MnistE-568}, where three learners, one from each distinct algorithm family, become oracles on three digits, 5, 6, and 8. In \textbf{AmpData}, one classifier is added as an oracle on two designated classes, being completely random anywhere else - complementing all other classifiers, which are random on those designated classes.

These setups evaluate whether DEEM can effectively identify experts, route relevant inputs to them and integrate their outputs—even under limited overlap with general knowledge. Further details and additional experiments are provided in Appendix~\ref{analysis-appendix}.

\begin{figure}[!t]
\centering
\includegraphics[width=1.\linewidth]{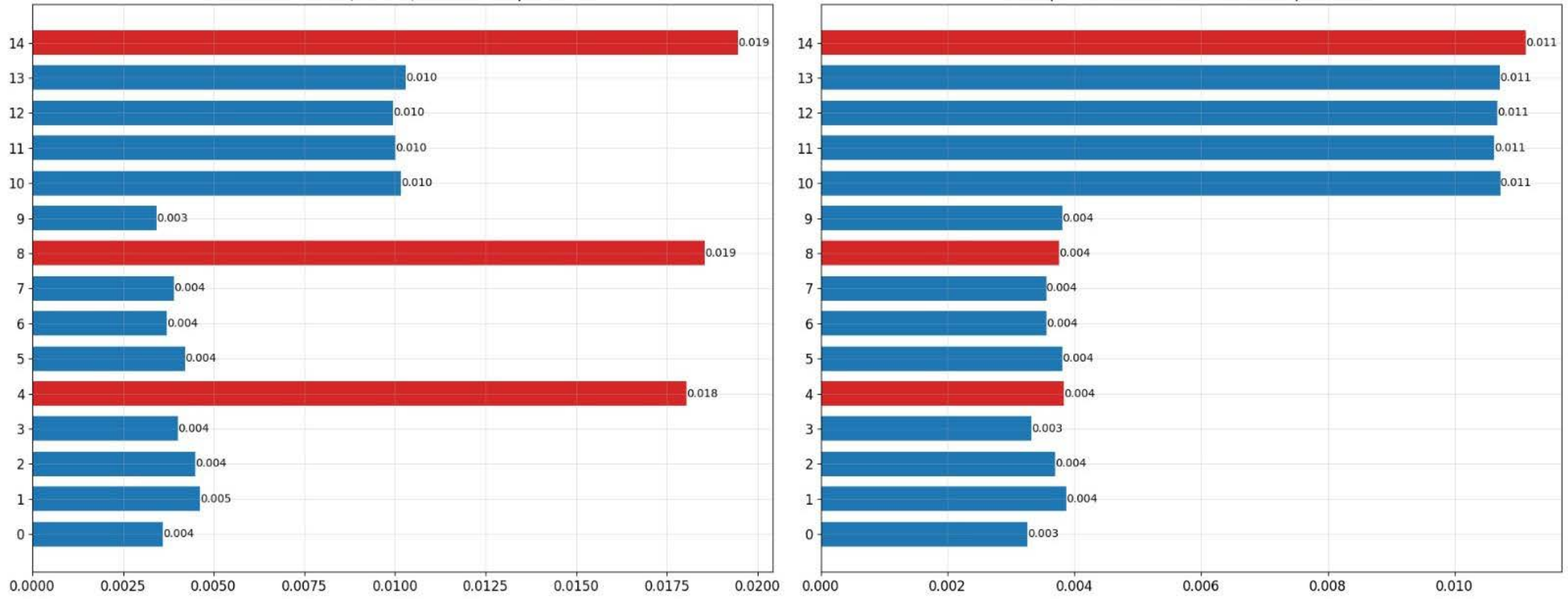}
\caption{Learner importance chart on MnistE-568. Red bars are the enhanced oracle learners. DEEM strongly favors the specialized on the expert, where they are relevant (left panel), while preserving their respective contribution on the rest of the data (right panel).}
\label{fig:moe}
\end{figure}

DEEM achieves best or second‑best accuracy on all portions of the datasets (Table~\ref{table:moe}). In AmpData, DEEM gains \textbf{7.34\%} over DS—where all other methods fail to harness the dataset’s oracle expertise, at a slight decrease in accuracy on the remaining data. DEEM also assigns 3–7× higher weights to subset experts (Figure~\ref{fig:moe}), showing it effectively synthesizes latent specializations.
\subsection{ImageNet Pretrained Ensemble}\label{subsec:imgnet}

To assess our method under harsh real-world large scale conditions, we created an ensemble of five pretrained models on the ImageNet validation set, where we filter out examples where all models unanimously agree. To our knowledge, this is the largest class-wise evaluation attempted in unsupervised ensemble learning, with two orders of magnitude more classes than previous benchmarks, expanding to $K=1000$.  Further details are available in~\cref{imagenet}.

Table~\ref{tab:imgnet-paper} shows DEEM achieves second best rankings, successfully scaling to beyond conventional class sizes, highlighting its robustness and relevance despite its sparse one-hot structure.
\begin{table}[!h]
    \centering
    \caption{ImageNet Accuracy}
    \label{tab:imgnet-paper}
     \resizebox{0.85\linewidth}{!}{%
    \smaller 
    \small
    \setlength{\tabcolsep}{4pt} 
    \renewcommand{\arraystretch}{0.9} 
    \begin{tabular}{cccc}
    \toprule
         BestClf&  MV&  L-SML&  DS\\
         \midrule[0.5pt]
         60.83&  56.41&  20.33&  56.24\\
 & & & \\
         \toprule
 LA$_{2pass}$& HLM& iRBM& DEEM
\\
 \midrule[0.5pt]
 \textbf{57.98}\unc{0.0}& 56.21& 57.37\unc{0.11}& \underline{57.47}\unc{0.05}\\
    \end{tabular}}
    
\end{table}
\section{CONCLUSIONS}
In this paper we presented DEEM, a novel flexible EBM model that enables unsupervised training and inference ensemble learning, with no dependence restrictions on classifiers. We lay out theoretical guarantees with empirical support for recovering the true posterior in the iRBM setting, and empirically demonstrate that multinomial layers reduce conditional dependencies. We benchmark DEEM performance compared to other baselines and showcase its properties given unique ensembling scenarios, including the ability to utilize leaners' expertise and accommodating large scale number of class sizes. We discuss limitations, scalability and ethic considerations in \autoref{considerations-appendix}. 
\subsubsection*{Acknowledgements}
We would like to thank Yilun Du and Léo Grinsztajn for their useful comments and observations. 
\clearpage
\bibliographystyle{abbrvnat}
\bibliography{ref}
\clearpage
\section*{Checklist}

The checklist follows the references. For each question, choose your answer from the three possible options: Yes, No, Not Applicable.  You are encouraged to include a justification to your answer, either by referencing the appropriate section of your paper or providing a brief inline description (1-2 sentences). 
Please do not modify the questions.  Note that the Checklist section does not count towards the page limit. Not including the checklist in the first submission won't result in desk rejection, although in such case we will ask you to upload it during the author response period and include it in camera ready (if accepted).

\textbf{In your paper, please delete this instructions block and only keep the Checklist section heading above along with the questions/answers below.}

\begin{enumerate}

  \item For all models and algorithms presented, check if you include:
  \begin{enumerate}
    \item A clear description of the mathematical setting, assumptions, algorithm, and/or model. Yes, in the main paper in~\cref{section:proposed-approach} and~\cref{fm-rbm-appendix,ci-proof-appendix,datasets-appendix}
    \item An analysis of the properties and complexity (time, space, sample size) of any algorithm. Yes, in the main paper in~\cref{section:proposed-approach} and relevant~\cref{datasets-appendix,additional-training-info-appendix}
    \item (Optional) Anonymized source code, with specification of all dependencies, including external libraries. Yes. Will be provided upon acceptance.
  \end{enumerate}

  \item For any theoretical claim, check if you include:
  \begin{enumerate}
    \item Statements of the full set of assumptions of all theoretical results. Yes. In~\cref{subsec:irbm,subsec:unsupervised_ensemble}
    \item Complete proofs of all theoretical results. Yes. Refer to~\cref{ci-proof-appendix,mle-proof-appendix}
    \item Clear explanations of any assumptions. Yes. please see aforementioned sections and appendices.     
  \end{enumerate}

  \item For all figures and tables that present empirical results, check if you include:
  \begin{enumerate}
    \item The code, data, and instructions needed to reproduce the main experimental results (either in the supplemental material or as a URL). Yes
    \item All the training details (e.g., data splits, hyperparameters, how they were chosen). Yes. Please see~\cref{additional-training-info-appendix}.
    \item A clear definition of the specific measure or statistics and error bars (e.g., with respect to the random seed after running experiments multiple times). Yes. Elaborated in main paper.
    \item A description of the computing infrastructure used. (e.g., type of GPUs, internal cluster, or cloud provider). Yes. \cref{additional-training-info-appendix}
  \end{enumerate}

  \item If you are using existing assets (e.g., code, data, models) or curating/releasing new assets, check if you include:
  \begin{enumerate}
    \item Citations of the creator If your work uses existing assets. Yes.
    \item The license information of the assets, if applicable. Yes. \cref{datasets-appendix}.
    \item New assets either in the supplemental material or as a URL, if applicable. Yes. Provided upon acceptance.
    \item Information about consent from data providers/curators. Not Applicable.
    \item Discussion of sensible content if applicable, e.g., personally identifiable information or offensive content. Not Applicable
  \end{enumerate}

  \item If you used crowdsourcing or conducted research with human subjects, check if you include:
  \begin{enumerate}
    \item The full text of instructions given to participants and screenshots. Not Applicable.
    \item Descriptions of potential participant risks, with links to Institutional Review Board (IRB) approvals if applicable. Not Applicable.
    \item The estimated hourly wage paid to participants and the total amount spent on participant compensation. Not Applicable.
  \end{enumerate}

\end{enumerate}

\onecolumn
\newpage
\appendix

\section{Fully Multinomial RBM}\label{fm-rbm-appendix}
We further describe the Fully-Multinomial Restricted Boltzmann Machine (hereby FM-RBM or simply RBM) in detail, writing its energy and derive its units' conditional probabilities.

Unless otherwise stated, capital letters denote sizes, and their lowercase letters are indices in range 1 to the capital letter. We denote $L,M$ as the visible and hidden multinomial unit sizes, respectively. Let $d_v,d_h$ be the RBM visible and hidden sizes, as in binary RBM.

$e_k$ is the one-hot vector with 1 at its index $k$. $e_k = \{e_1 \dots e_K\}$ where K is the number of classes. In our derivation it will be L or M for $v$ or $h$, respectively.

The energy of the FM-FBM is:
\[
E(v, h) = - (\sum_{l}\sum_{i} a^l_{i} v^l_{i} + \sum_{m}\sum_{j} b^m_{j} h^m_{j} + \sum_{l}\sum_{m}\sum_{i}\sum_{j} w^{lm}_{ij} v^l_{i} h^m_{j})
\]
Or in its vectorized form:
\begin{align}
E(v, h) &= -\sum_{l} <a^l,v^l> - \sum_{m} <b^m,h^m> - \sum_{l}\sum_{m} <<w^{lm},v^l>,h^m> \\ &=-(\sum_{l} <a^l,v^l> + \sum_{m} <b^m,h^m> + \sum_{l}\sum_{m} <<w^{lm},v^l>,h^m>) \\
&= - \left(\sum_{l} {a^l}^T v^l + \sum_{m} {b^m}^T h^m \sum_{l,m} {v^l}^T W^{l,m} {h^m}\right)
\end{align}
This follows as an expansion to support multinomial input and output, as first proposed in \citep{salakhutdinov2007collaborative} for the Multinomial-Binary RBM.

\section{FM-RBM Visible Conditional Probability }\label{fm-rbm-ci-appendix}
Recall that $v_i$ is a multinomial vector with $L$ elements. The only allowed configurations for a multinomial vector are the one-hots defined by $\mathbf{e}_1 \dots \mathbf{e}_L$. which we will denote as $\mathbf{e}_l$ for the one-hot vector with 1 at the $l$-th index.

{\large
\begin{align}
p(v_i=\mathbf{e}_l|h) &= \frac{p(v_i=\mathbf{e}_l,h)}{p(h)} \\
&= \frac{\sum_{\{v|v_{i'}=\mathbf{e}_l\}} p(v_{i'},h)}{\sum_v p(v_{i'},h)} \\
&= \frac{\sum_{\{v|v_{i'}=\mathbf{e}_l\}} e^{-E(v,h)}/Z}{\sum_v e^{-E(v,h)}/Z} \\
&= \frac{\sum_{\{v|v_{i'}=\mathbf{e}_l\}} e^{-E(v,h)}}{\sum_{\{v|v_{i'}=\mathbf{e}_l\}} e^{-E(v,h)} + \sum_{\{v|v_{i'}\ne \mathbf{e}_l\}} e^{-E(v,h)}} \label{eq:fmrbm-vis-condprob-init} 
\end{align}
}%

Note that we can write 
\[\sum_v e^{-E(v,h)} = \sum_{l=1}^L \sum_{v|v_{i}=\mathbf{e}_l} e^{-E(v,h)}\]
for any $v_i$, as we defined its allowed values earlier.

Now let us look at the energy in numerator term from \autoref{eq:fmrbm-vis-condprob-init}, for the specific $\mathbf{e}_l$:

{\large
\begin{align}
\sum_{\{v|v_{i'}=\mathbf{e}_l\}} {-E(v,h)} &= \sum_{l=k} \sum_{i=r} a^l_i g^l_i + \sum_m \sum_j b^m_j h^m_j + \sum_{l=k} \sum_m \sum_{i=r} \sum_j g^l_i w^{lm}_{ij} h^m_j \\
&=  a^k_r \cdot 1 + \sum_m \sum_j b^m_j h^m_j + \sum_{l=k} \sum_m \sum_{i=r} \sum_j 1 \cdot w^{km}_{rj} h^m_j \\ &= a^k_r + \sum_{m,j} b^m_j h^m_j + \sum_{l=k,m,i=r,j} w^{km}_{rj} h^m_j \\ &= a^k_r + \sum_{m,j} b^m_j h^m_j + \sum_{m,j} w^{km}_{rj} h^m_j 
\end{align}
}%
Note that the term $\sum_{m,j} b^m_j h^m_j$ doesn't depend on the values of  $l$ or $i$,

Therefore:
{
\begin{align}
p(v_i=\mathbf{e}_k|h) &= \frac{\exp(a^k_r + \sum_{m,j} b^m_j h^m_j + \sum_{m,j} w^{km}_{rj} h^m_j )}{\exp(a^k_r + \sum_{m,j} b^m_j h^m_j +  \sum_{m,j} w^{km}_{rj} h^m_j + \sum_{l\ne k, i=r} a^l_r + \sum_{m,j} b^m_j h^m_j + \sum_{l \ne k, m, i=r, j} w^{lm}_{rj} h^m_j)} \\
&= \frac{\exp(a^k_r + \sum_{m,j} w^{km}_{rj} h^m_j) }{\exp(a^k_r + \sum_{m,j} w^{km}_{rj} h^m_j + \sum_{l\ne k, i=r} a^l_r + \sum_{l \ne k, m, i=r, j} w^{lm}_{rj} h^m_j)} \\
\end{align}
}%

Which ultimately gives the softmax probability:
\[
p(v_i=\mathbf{e}_k|h) = \frac{e^{a^{k}_i + \sum_{m}\sum_{j} w^{km}_{ij} h^m_{j}}}{\sum_{l=1}^{L} e^{a^{l}_i + \sum_{m}\sum_{j} w^{lm}_{ij} h^m_{j}}}
\]
A derivation in similar fashion can be done for the hidden conditional probability $p(h_j=k|v)$ as well.

\section{Proof of Lemma 1}\label{ci-proof-appendix}
\begin{proof} We will define $\theta$ so for every $x,y$, $p_\theta(X_i = l|Y = m) = p_\lambda(V_i = l|H = m)$, and $p_\theta(Y = y) = p_\lambda(H = y)$. For completeness, in this proof we denote $L$ as the visible multinomial dimension, and $M$ as the hidden multinomial dimension. The weight matrix $W$ has dimensions $(L,M,d,1)$, the visible bias $a$ has dimensions of $(L,d)$ and the hidden bias $b$ dimension is $M\times1$. 

Recall that:
\[
p_\lambda(V_i=\mathbf{e}_{l}|H_j=\mathbf{e}_{m}) = \frac{e^{a^{l}_i + \sum_{m}\sum_{j} w^{lm}_{ij} h^m_{j}}}{\sum_{l'=1}^{L} e^{a^{l'}_i + \sum_{m}\sum_{j} w^{l'm}_{ij} h^m_{j}}} 
\]
Because our hidden dimension is 1, we can remove the summation over $j$. We also denote $H \equiv H_1$. Moreover, note that $H$ is the one-hot vector $\mathbf{e_m}$. This let us reduce the equation to:
\[
p_\lambda(V_i=\mathbf{e}_{l}|H=\mathbf{e}_{m}) = \frac{e^{a^{l}_i + w^{lm}_{i}}}{\sum_{l'=1}^{L} e^{a^{l'}_i + w^{l'm}_{i}}} 
\]
Therefore we define: 
\[
\psi_{ilm} \equiv \sigma(\mathbf{z})_{ilm} 
\]
Where $\sigma$ is the softmax function, $z_{ilm}=(a_i^l + w_i^{lm})$, and $\mathbf{z} = (z_{i1m},\dots,z_{iLm}) \in \mathbb{R}^L$.

Note that for iRBM, for any combination of indices $(l, m) \in \mathcal{L} \times \mathcal{M}$ such that either $l = 1$ or $m = 1$ (The first visible or hidden index), we set the parameters $\left(a_i^l, w_i^{lm}\right)$ to be some fixed scalar values, denoted as $\bar{a}_i$ and $\bar{w}_i^{lm}$ \footnote{In our paper they equal 0 or 1}. This ultimately make the effective parameters of $W$ and $a$ be $(L-1)\times (M-1) \times d$ and $(L-1)\times d$, respectively. Since in our case, $K = L = M$, we can simply write $d\times (K-1)^2$ and $d\times (K-1)$.

Notice that the map from the effective parameters $(a_i^l,w_i^{lm})$ to $\psi_{ilm}$ is injective, since there are $d(K-1) K$ unknowns and `the same number of linearly independent equations.

Assume towards a contradiction, that the map from the effective parameters $(a_i^l,w_i^{lm})$ to $\psi_{ilm}$ is not injective. Then there exist two distinct pairs $(a_i^l,w_i^{lm})\neq(\tilde{a}_i^l,\tilde{w}_i^{lm})$ such that for some set of indices $(i,l,m)$, 
\[
 z_{ilm} = (a_i^l + w_i^{lm})=(\tilde{a}_i^l + \tilde{w}_i^{lm})
\]
Since the $w_i^{lm}$ are unique to each $\psi_{ilm}$, it must be that any change in parameters comes from altering $a_i^l$ as well.
However, notice that modifying any $a_i^l$  necessarily inflict change on $w_i^{l1}$ for $z_{il1}$ (to keep the same sum). 
This contradicts Definition \autoref{fm-irbm}, in which $w_i^{lm}$ for $m=1$ is a constant.
Therefore, the map $(a_i^l,w_i^{lm})$ to $\psi_{ilm}$ is injective.

By FM-RBM definition, recall that:
\begin{align*}
    p_\lambda(H=\mathbf{e}_t) &= \frac{\sum_{v\in \{\mathbf{e}_1,\dots,\mathbf{e}_L\}^d} e^{-E(v,\mathbf{e}_t)}}
{\sum_{v\in \{\mathbf{e}_1,\dots,\mathbf{e}_L\}^d,h\in \{\mathbf{e}_1,\dots,\mathbf{e}_M\}} e^{-E(v,h)}} \\
&= \frac{\sum_v e^{\sum_{l} {a^l}^T v_l + \sum_{m} {b^m}^T h_m + \sum_{l,m} v^l W^{l,m} {h^m}^T}}
{\sum_{v,h} e^{\sum_{l} {a^l}^T v_l + \sum_{m} {b^m}^T h_m + \sum_{l,m} v^l W^{l,m} {h^m}^T}}
\end{align*}
For brevity, we write $\sum_{v\in \{\mathbf{e}_1,\dots,\mathbf{e}_L\}^d}$ as  $\sum_v$, and $\sum_{h\in \{\mathbf{e}_1,\dots,\mathbf{e}_M\}}$ as $\sum_h$.

Since $H=\mathbf{e_t}$, the hidden term $\sum_{m} {b^m}^T h_m$ simply reduces to $b^t$ (As it only equals 1 in the $t$-th dimension), while visible term $\sum_{l} {a^l}^T v_l$ can be factorized out. Thus we set:
\[
\pi_t \equiv \frac{\sum_v e^{b^t + \sum_{l} v^l W^{l,t}}}
{\sum_{v,h} e^{b^m + \sum_{l} v^l W^{l,m}}}
\]
To see the map of $\lambda \longmapsto \theta$ is 1:1, if we open the summation over $h$, we can rearrange this equation and get:
\begin{align*}
    &\pi_t (\sum_v e^{b^t + \sum_{l} v^l W^{l,t}} + \sum_{v,h\neq e_t} e^{b^m + \sum_{l} v^l W^{l,m}}) = \sum_v e^{b^t + \sum_{l} v^l W^{l,t}} \\
    &\Rightarrow \pi_t \sum_{v,h\neq e_t} e^{b^m + \sum_{l} v^l W^{l,m}} = (1 - \pi_t) \sum_v e^{b^t + \sum_{l} v^l W^{l,t}} \\
    &\Rightarrow e^{b^t} = \frac{\pi_t}{1 - \pi_t} \cdot \frac{\sum_{v,h\neq e_t} e^{b^m + \sum_{l} v^l W^{l,m}}}
    {\sum_v e^{b^t + \sum_{l} v^l W^{l,t}}}
\end{align*}
Therefore, given $(W,a)$, $\pi_t$ can be uniquely determined by $b^t$. As per iRBM definition, the effective parameters count of $b$ is $d_h * (M-1)$, or just $(K-1)$ in our case. This means that the map from $b^t$ to $\pi^t$ is injective, as there are $(K-1)$ linearly independent equations for $\pi^t$ as well.

To complete the proof, notice that the total parameters of both model are equal. For the conditional independence model, there are a total of $d(K-1)K$ parameters for the conditionals, and $K-1$ for the priors. This yields the identity:
\[
d(K-1)^2 + d(K-1) + (K-1) = d(K-1)K + (K-1) 
\]
In total, each model has exactly $(dK+1)(K-1)$ parameters. 
 
As a result, the map $\lambda \longmapsto \theta$ is injective. An identical argument shows that the map $\theta \longmapsto \lambda$ is injective as well, therefore the map $\lambda \longmapsto \theta$ is a bijection.
\end{proof}
\section{Proof of Corollary 2} \label{mle-proof-appendix}
\begin{proof} Given the conditions outlined in Corollary \autoref{corollary:fm-irbm-identifiable} under which the iRBM is identifiable, and as $p_\lambda(H=y|V=x)$ is continuous in $\theta$, by the consistency property of MLE, we get:
\[
\lim_{n\to\infty}{\hat{\lambda}_\text{MLE} = \lambda}
\]
and
\[
p_{\hat{\lambda}_\text{MLE}}(H=y|V=x) \to p_{\lambda}(H=y|V=x)
\]
Under the map $\lambda \mapsto \theta$,  we acquire
\[
p_{\lambda}(H=y|V=x) \to p_{\theta}(H=y|V=x)
\]
which completes the proof. \end{proof}

\subsection{MLE Consistency}
We appeal to Wald’s Consistency Theorem to guarantee that the Maximum Likelihood Estimator $\hat{\lambda}_{MLE}$ converges to the true parameter $\lambda_0$. Our iRBM framework satisfies the required regularity conditions:
\begin{itemize}
    \item Compactness: The parameter space (probabilities and weights) is a compact subset of Euclidean space.
    \item Identifiability: The unique maximizer condition is satisfied; The global identifiability of the iRBM is rigorously proved in  \cref{lem:ci-rbm} and~\cref{remark:iden}.
    \item Continuity: The log-likelihood of the iRBM (an exponential family model) is continuous with respect to $\lambda$.
    \item Uniform Convergence: The energy function is regular, ensuring that the Uniform Law of Large Numbers applies. Since these conditions hold, the estimator is consistent, ensuring the posterior convergence stated in~\cref{corollary:posterior}.
\end{itemize}
Therefore, the MLE is consistent. 
\section{Additional Background}\label{background-appendix}

\subsection{Energy-Based Models}
Energy-Based Models (EBMs) are a class of probability models, that associate an energy scalar $E_w(x)$ to each configuration of $x$ w.r.t. to learnable parameters $w$. For a given input $x$, its probability is defined as the Boltzmann distribution over all possible configurations:
\begin{equation}
p_w(x) = \frac{\exp(-E_w(x))}{Z(w)}    
\end{equation}
where $Z(w)=\int \exp(-E_w(x))\,dx$ is the normalization constant.

The probability $p_w(x)$ is proportional to the exponential of the negative energy, ensuring that configurations with lower energy have a higher probability.
The gradient of the maximum log-likelihood is given by:
\begin{equation}
\frac{\partial}{\partial w} \log p_w(x) = - \frac{\partial}{\partial w} E_w(x) + \mathbb{E}_{x'\sim p_w}\left[\frac{\partial}{\partial w} E_w(x')\right]
\label{eq:grad-term-ebm}
\end{equation}
The first term can be easily computed, but the computation of second term (which refers to the configurations sampled from the EBM distribution), is intractable in most cases, so the gradient can only be approximately estimated by sampling from $p_w$. One common way is by constructing a Markov chain.

In the context of learning, \eqref{eq:grad-term-ebm} can be interpreted as adjusting the parameters $w$ of the model to minimize the energy of the observed data points while maximizing the energy of the data points currently sampled from the model, ultimately associating high energy to unobserved points. 

\subsection{Restricted Boltzmann Machine}
Restricted Boltzmann machine (RBM) is a bipartite graph generative stochastic EBM, with a set $V$ of $d_v$ visible units, and a set $H$ of $d_h$ hidden units, and each unit takes binary values. $V$ and $H$ are arranged in two layers, fully connected to one another, with no connections between units of the same layer.

An RBM is parameterized by a weight matrix $W$ of size $(d_v,d_h)$, in addition to a visible bias $a$ of size $d_v$ and a hidden bias $b$ of size $d_h$. Given these parameters, the energy function of the RBM for a given configuration of $v$ and $h$ is defined as:
\begin{equation}
   E(v,h) = - \left({a}^T v + {b}^Th  + {v}^T W {h}\right)
\label{eq:rbm-energy}
\end{equation}
With the joint probability distribution:
\[
p(v,h) = \frac{e^{-E(v,h)}}{\sum_{v,h}e^{-E(v,h)}}
\]
The target of training an RBM is to learn a probability distribution over a set of data, mostly to generate new similar data. The visible units represent observed data points and the hidden units correspond to a latent representation we learn. Each hidden unit can be interpreted as a learner function that captures some hidden dependency between observed features~\citep{fischer2014training,montúfar2014number}.

\section{FM-RBM Input Example}\label{fm-rbm-example-appendix}
Consider a FM-RBM as described in \autoref{section:background} with $d=4$ classifiers and $K=3$ classes. 
Given an example input $[1, 3, 2, 1]$, it is transformed into a 2D one-hot matrix $x_i \in \{0,1\}^{K \times d}$:
\begin{equation}
x_i = \begin{bmatrix}
1 & 0 & 0 & 1 \\
0 & 0 & 1 & 0 \\
0 & 1 & 0 & 0 \\
\end{bmatrix}
\end{equation}
where each column is a one-hot vector corresponding to the classifier's prediction.
For the forward pass, the FM-RBM weight tensor $W \in \mathbb{R}^{K \times K \times d_v \times d_h}$ is applied to $x_i$, adding the hidden bias $b \in \mathbb{R}^{K \times d_h}$. This effectively computes:
\begin{equation}
z^{m}_{j} = \sum_{l=1}^K \sum_{i=1}^{d_v} W^{lm}_{ij} x^{l}_{i} + b^{m}_{j}
\end{equation}
for each $m \in {1,\ldots,K}$ and $j \in {1,\ldots,d_h}$. Where $z^{m}_{j}$ is the logit (Before applying softmax on the $m$-th dimension).

\subsection{Identifiable RBM Visualization}
\label{irbm-visualization}
In essence, the Fully Multinomial Identifiable RBM (thereby iRBM) only modification to the standard FM-RBM, is fixing the first coefficients of every visible and hidden multinomial units of the model. 

This modification lets $K-1$ degrees of freedom for every probability vector of $K$ classes, as one can notice that for a probability vector of size $K$, the last coefficient is always the remainder of one minus the sum of all other probabilities (As can be seen in Remark \ref{ci-model-remark}, for example). This is what allow us to uniquely determine the parameters that produced the values. 

\begin{figure}[h]
    \centering
    \includegraphics[width=0.75\linewidth]{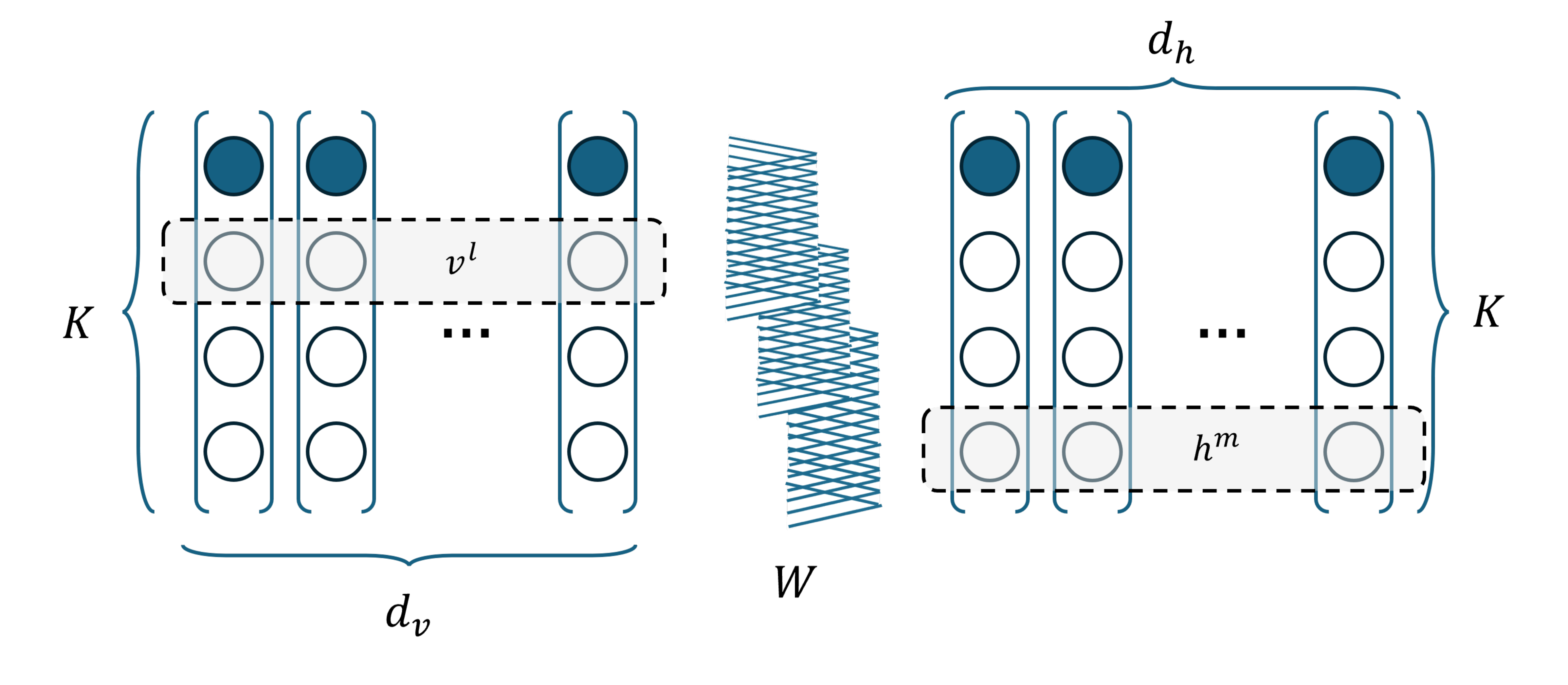}
    \caption{Fully-Multinomial Identifiable RBM (iRBM). The filled units are the fixed coefficients of the model.}
    \label{fig:irbm-fixed-visualization}
\end{figure}

\section{Additional Training and Inference Information} \label{additional-training-info-appendix}
Here we lay out additional training and inference information for using DEEM in practice.
\subsection{Initalization Scheme}
In our implementation we initialize the Multinomial layers to the identity function:
\[
w^{lm}_{ij} = \begin{cases}
1 & \text{if } l = m \text{ and } i = j \\
0 & \text{otherwise}
\end{cases}
\]
We add $\mathcal{N}(0, \sigma^2)$ noise to both weights and bias, where $\sigma = 0.005$ by default.

The iRBM is initialized to output the majority vote decision:
\[
w^{lm}_{i} = \begin{cases}
1 & \text{if } l = m \\
0 & \text{otherwise}
\end{cases}
\]
Then, for all zeroed weights and all biases:
\[
w^{lm}_{i} = \epsilon, \quad a^l_i = \epsilon, \quad b^m = \epsilon
\]
where $\epsilon \sim \mathcal{N}(0, \sigma^2)$ and $\sigma = 0.01$ by default.

This lets us begin training when every learner is unfavorable and has equal amount of contribution to the meta prediction. This initialization returns the majority vote, as the $m$-th multinomial logit accumulates 1 only when $l=m$, meaning the learner predicted the class $m$. Since we sum over $d$ and then apply the softmax function, the highest probability will be the class that got the most predictions.
\subsection{Model Distribution Sampling Method}
Traditional RBM gradient approximation methods like Contrastive Divergence \citep{Carreira2005cd} or PCD \citep{tieleman2008pcd} rely on MCMC sampling between visible and hidden layers, limiting training to a layer-wise approach. To maintain end-to-end flexibility, we adopt an alternative sampling strategy.

We employ the deep Langevin proposal (DLP) sampling method \citep{zhang2022langevinlike} for our energy-based model. This scalable, parameter-free approach constructs a proposal distribution over possible local updates based on likelihood gradients with respect its discrete input. In DEEM, these updates encompass all $K$ classes for each learner. Refer to \autoref{dlp-appendix} for the pseudo-code. 

\subsection{DLP Sampler}
\label{dlp-appendix}
The pseudo code for our sampling methods follows the recent work presented by \cite{zhang2022langevinlike}, which suggest a scalable gradient-based sampling approach for discrete distributions. The pseudo-code for the algorithm is:
\begin{algorithm}
\caption{Samplers with Discrete Langevin Proposal (DULA and DMALA)}
\label{algorithm:dula_dmala}
\SetAlgoLined
\small
\DontPrintSemicolon
\KwIn{$\alpha$: step size}
\KwOut{$\lambda^{(k)}$: samples}
\SetKwProg{Proc}{Procedure}{}{}
\Proc{Sample With DLP}{}{
Initialize $\lambda$\;
\While{convergence criterion not met}{
    \For{$i \gets 1$ \KwTo $d_\lambda$}{ \tcp{Done in parallel}
        Construct $q_i(\cdot|\lambda)$ by \eqref{eq:dlp-q} \;
        Sample $\lambda_i \sim q_i(\cdot|\lambda)$\;
    }
    \tcp{MH Step}
    Compute $q(\lambda'|\lambda) = \prod_i q_i(\lambda_i'|\lambda)$\;
    Compute $q(\lambda|\lambda') = \prod_i q_i(\lambda|\lambda'_i)$\;
    Set $\lambda \leftarrow \lambda'$ with probability given by \eqref{eq:dlp-mh} \;
}
\Return $\lambda^{(k)}$
}
\end{algorithm}

Where $q_i(\lambda'|\lambda)$ is a categorical distribution:
\begin{equation}
q_i(\lambda_i) = \text{Categorical}\left(\text{Softmax}\left(\frac{1}{2}\nabla U(\lambda)_i^T(\lambda_i' - \lambda_i) - \frac{\lVert \lambda'_i - \lambda_i\rVert_2^2}{2\alpha}\right)\right),
\label{eq:dlp-q}
\end{equation}
 and $\lambda_i,\lambda_i'$ are one-hot vectors.

We use their DMALA algorithm, which adds a Metropolis-Hastings correction to the suggested proposal. The MH step equation is given by:
\begin{equation}
\min \left(1, \exp(U(\lambda') - U(\lambda)) \frac{q(\lambda|\lambda')}{q(\lambda'|\lambda)}\right)
\label{eq:dlp-mh}
\end{equation}
In our experiments, we found it to perform slightly better than the gibbs based approach suggested by \cite{grathwohl2021oops}, and with more sampling speed. Another recent Langevin-based method by \cite{du2019implicit} performed well on some seeds, but we found DLP to be more consistent and robust to hyperparameters choices, likely since it is tailored for discrete inputs.

\subsection{Choosing Hyperparameters and Training Insights}
\label{hyperparameters-appendix}
Due to the energy-based nature of our model, and especially its training procedure, which involves maximizing a proxy to the gradient (As pointed in Remark \autoref{remark:achieving-mle}), careful attention to its chosen hyperparameters and model configuration is required to avoid unwanted behavior. Hyperparameters that are too low will result in long or non-existent training, while hyperparameters that control the learning too much might interfere the delicate two-part energy training dynamic, possibly making data examples' energy go up or retrieving poor model negative samples, causing training instability or diversion.

Another issue we encountered repeatedly during our initial research is that for some configurations, like those with an inadequate sampler or with a learning rate that is too strong, resulted in a phenomenon we call 'dead units', where during training the model collapses and maps all data to a smaller subset of its multinomial output units.  Since our method is fully unsupervised, and we do not have any other data information, we can not tell if this is a wanted behavior for a given dataset (As some data might be inherently unbalanced), which leads us to derive training insights from other metrics. 

Wherever applicable, we put an emphasis in our trials on figuring good hyperparameters and configurations that need less compute, in order to have a solid training foundation that is easier and faster to run for as many applications.

We found these tips to help boost the model stability significantly:
\begin{itemize}
    \item A discrete sampler that allows for quality model samples. We were able to create performant models using other EBM sampler method, but as pointed in the aforementioned section, they were less robust. For our DLP sampler, we found that a small number of sampler steps (number of times we run a sample through the model repeatedly to acquire a truer model sample) is sufficient. We have used 5, trying values up to 100, but the reader is advised to experiment with more sampler steps given the resources.
    \item Our trials suggest regularization is unnecessary for training DEEM successfully. Although it is usually advised, using too big of a regularizer can cause dead units to appear, which overshadows potential benefits. If you choose to use regularization, consider L1 over L2.
    \item Bigger batch sizes are significantly better than small ones. It is possible to train using as low as 32 examples per batch, but results are often less stable. We opt for a batch size of at least 1024 examples universally, with more if memory and compute allows.
    \item We saw that SGD momentum boosts training in the beginning, but since we do not have a fine-grained metric to elucidate where we reach convergence, we ultimately chose not to use it. If you do choose to use momentum, values up to 0.5 are enough.
    \item The right convergence criterion is one that allows the model to stabilize after possible energy difference dips (more on that in the section below). However, in practice, it has shown that with a proper learning rate, top performance can be achieved sufficiently within a range of 50-150 epochs.
    \item In our experimentation, one multinomial layer already should suffice to capture a reliable amount of the benefits of the disentanglement and reduction of conditional independence between learners. However, it is worth noting (as discussed in the appendices below) that most benchmarks in this field are of small to medium size datasets, and wherever the dataset was bigger in size or class number (Like MnistE, Tree3k or ImageNet), additional layers have been helpful in harvesting more of the aforementioned benefits, with some diminishing returns. As a rule of thumb, one can begin with one multinomial layer, and decide to add more given the other metrics or phenomena (like the MI scores or 'dead units') are still in line. The other real-world consideration might be the added compute time overhead for propagating the gradient through the added layers.
    \item We have found that initializing with a small gaussian noise and increasing the impact of the MV initialization (i.e. setting $w_{ij}^{11}=2$ ) improves stability and robustness to a degree. We only included this in the appendix in order to avoid confusion, as the base MV initialization is stable enough. 
\end{itemize}

\subsubsection{Choosing the learning rate}
As for the learning rate, we found that using large values often resulted in great overall performance, but opened the training process for instability and the dead units phenomenon that badly hurts accuracy, which is hard to spot and stop mid-training. However, a learning rate that is small but can give subpar results, where too small of a learning rate causes model to stay stagnant. Therefore, a thoughtful choice is needed for choosing a learning rate for a new given dataset, making sure to balance the trade-off between stability and accuracy.

In order to understand how the learning rate behaves, and to determine what could be a good candidate for a given dataset, we run an analysis test using a set of potential learning rate values for one fixed set of initial weights initialization. In our analysis, we run every configuration for a fixed number of epochs, and look at one main metric and two complementary metrics, computed on the whole training dataset:
\begin{itemize}
    \item Energy difference: The difference between two terms, the mean (positive) data points energy and the (negative) mean model samples energy. 
    \item Positive energy: The mean positive energy, the first term of energy difference.
    \item Negative energy: The mean negative energy, the second term of energy difference.
\end{itemize}

\begin{figure}[h]
    \centering
    \includegraphics[width=0.5\linewidth]{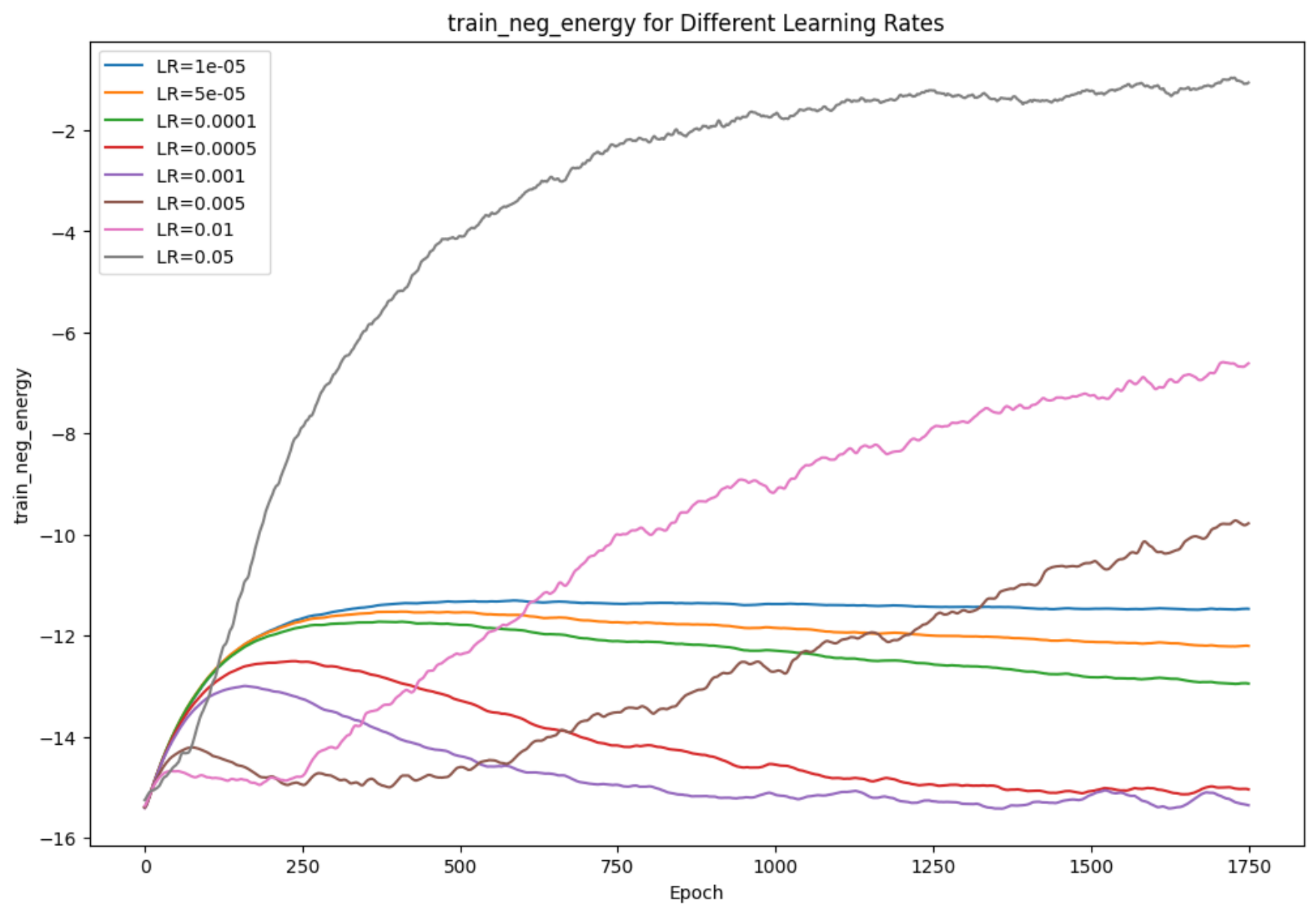}
    \caption{GesturePhsm negative energy graph for each learning rate. It can be seen that the first 3 biggest learning rates in this example increase over time, which implies these learning rate are unstable at that current hyperparameter setup.}
    \label{fig:phsm-neg-energy}
\end{figure}
Since our model is an energy-based model, we can expect to find the positive energy to not increase, meaning our data samples become more probable as model samples. We also expect to see similar behavior for the negative model samples, which means the samples we acquire using our sampler, correctly represent high probability model distribution samples. Seeing curves that behave otherwise can suggest the model training is unable to provide high quality model samples or learn the data distribution properly.

Given the two complementary metrics behave as expected, we experimentally found the best runs to be those which their energy difference metric to show a 'dip' at the beginning of training, where the difference value drops and increases, then relatively convergence to a value for as long as possible without dropping again, creating a U or a V shaped curve with a long tail. Values that dropped suggested model instability especially for higher learning rates. Small portion of datasets output a mirror image of these phenomena, where all curves start high and decrease, in that case the inverse of all guidelines apply.

In summary, we choose the promising learning rates based on the following steps:
\begin{itemize}
    \item Look at the positive and negative energy curves, and filter out LRs that their curves increase, see Figure~\ref{fig:phsm-neg-energy} for example.
    \item Filter out energy difference curves that either explode down or up (Including ones that show a 'dip', but then fall or rise from the tail). Usually, curve values stay within 2-2.5 times their initial value margin. 
    \item From all remaining LR curves, that their energy difference curve has a 'dip' and a tail, it is preferred to choose the smallest learning rate, for stability.
    \item As a general rule of thumb, our goal in maximizing the energy loss is to have zero or close-to-zero difference between the averaged energy values of the positive and negative examples, and so trials that approached zero tend to be better than one that did not. However, this is not granular and it is not guaranteed that the curve which is closest to zero has indeed the highest score.
\end{itemize}

\begin{figure}[h]
    \centering
    \includegraphics[width=0.5\linewidth]{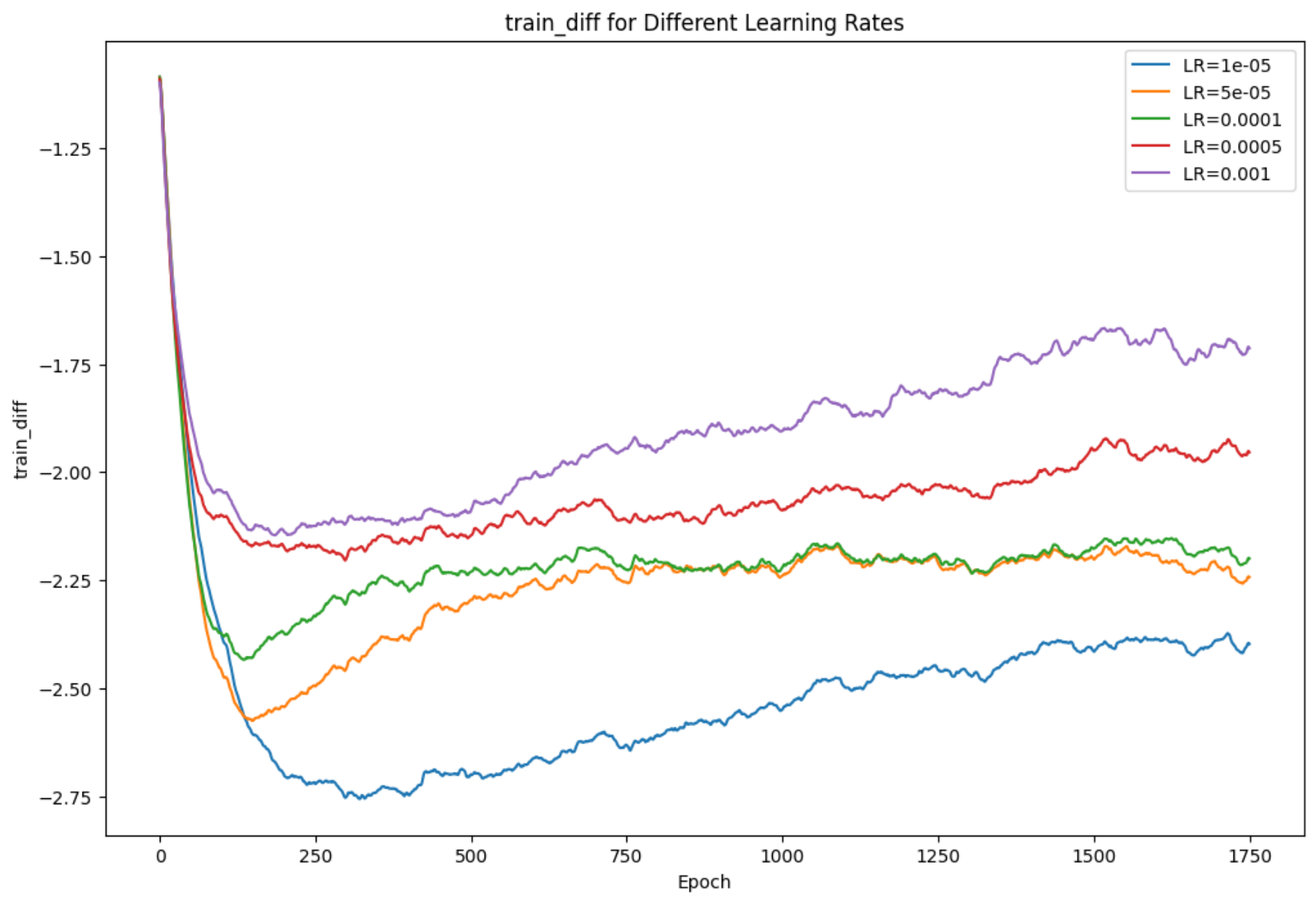}
    \caption{Tree3k energy difference for each learning rate, after filtering exploded values. In this plot, no learning rate was able to rise completely from the 'dip' and approach zero. However, 5e-5 and 1e-4 (orange and green lines) are potent due to their more aggressive 'dip', and the long steady, not increasing tail that follows.}
    \label{fig:tree3k-energy-diff}
\end{figure}
Notes:
\begin{itemize}
    \item Some energy difference curves might seem to converge to a tail in the future or have a wide 'dip'. These LRs might also be suited for the problem, but come second after ones that posses the aforementioned structure, as it is unknown whether they can actually reach convergence. In some cases, such energy curves might be favorable, given their individual positive and negative curves are more stable than the ones with the clearer 'dip'.
    \item For some plots, it is easier to spot the trends using smoothing. We use a simple exponential moving average with alpha values ranging from 0.7 to 0.99. 
\end{itemize}

Once a learning rate and other hyperparameters have been chosen, one can remove less successful runs based on the stability of their energy difference curve and the same guidelines above.

We employed an automated tool monitoring the energy difference metrics between different initializations. The yieled results were then manually analyzed to extract useful guidelines for choosing and handling the various hyperparamters, and understading how they behave in the training regime. We publish this as part of our codebase, in order to provide practitioners with an automated flow for choosing hyperparameters for everyday tasks,

\subsection{Running time and compute resources}
Table \autoref{tab:all_methods_comparison} presents the running time comparison for every dataset. It can be seen that regardless of the deep-learning infrastructure and the energy-based training, DEEM is still competitive with other classical baselines, and stays within the same order of magnitude, with a small overhead by absolute means. 

All experiments were conducted using one NVIDIA L40S GPU; however, the method is lightweight and can be executed on significantly slower hardware without issues. The memory footprint is minimal, requiring only about 0.5 GB of VRAM on average, making the approach accessible even on modest GPU setups. We believe this ensures reproducibility and ease of experimentation across a wide range of computational environments. For more details regarding scalability concerns please refer to \autoref{considerations-appendix}.
\begin{table}[h]
\centering
\caption{Comparison of methods running time across datasets (in seconds). Values are mean~$\pm$~std or raw times (lower is better).}
\label{tab:all_methods_comparison}
\resizebox{\textwidth}{!}{%
\begin{tabular}{lcccccccc}
\toprule
\textbf{Method} & Tree3k & MnistE & Csgo & Petfinder & MicroAgg2& EyeMovem& Arti Chars & Gesture Phsm \\
\midrule
MV & 0.64 & 1.11 & 0.05 & 0.30 & 0.54 & 0.25 & 0.24 & 0.29 \\
L-SML & 1.45 & 4.50 & 0.06 & 1.34 & 1.71 & 1.08 & 3.65 & 1.75 \\
DS & 5.68 & 53.03 & 1.43 & 8.04 & 11.29 & 3.88 & 17.25 & 14.21 \\
LA & \textless1s & \textless1s & \textless1s & \textless1s & \textless1s & \textless1s & \textless1s & \textless1s \\
EBCC & 
3.37$\pm$1.50 & 
31.64$\pm$4.52 & 
0.12$\pm$0.05 & 
8.58$\pm$0.81 & 
15.19$\pm$0.10 & 
2.63$\pm$0.75 & 
7.47$\pm$3.09 & 
6.06$\pm$1.28 \\
HLM & 1.02 & 2.83 & 0.03 & 0.47 & 0.55 & 0.30 & 0.60 & 0.87 \\
\midrule[0.5pt]
DEEM & 
16.12$\pm$0.43 & 
42.68$\pm$0.81 & 
7.35$\pm$0.68 & 
21.03$\pm$0.54 & 
15.02$\pm$0.58 & 
18.94$\pm$0.51 & 
44.75$\pm$0.70 & 
19.09$\pm$0.75 \\
\bottomrule
\end{tabular}
}
\end{table}

\subsection{Use of soft-labels}
In our paper, we focused on hard labels to align with the classic unsupervised ensemble and crowdsourcing literature and benchmarks. In many real-world black-box scenarios, one only has access to discrete decisions (e.g., human annotators, sensors, API calls), not internal probabilities.

However, our architecture is actually not constrained to hard labels, unlike some of the other methods. As we utilize one-hot encoding for the input of DEEM, and this structure is kept from the input through the iRBM component, the model can inherently accept continuous vectors (soft labels) from the probability simplex, with no further adjustments. This also allows to combine both soft and hard labels in one example if needed.

The following experiment shows the accuracy of DEEM trained on a new PetFinder ensemble dataset with adjusted classifiers trained on a different seed, using the classifiers' soft labels as input. 

\begin{table}[!h]
    \centering
    \caption{Accuracy of the revised PetFinder ensemble dataset}
    \begin{tabular}{cccc}\toprule
         BestClf&  AvgClf&  MV& DEEM\\\midrule
         94.29&  71.03\unc{17.85}&  82.24& 84.09\unc{0.29}\\ \bottomrule
    \end{tabular}
    \label{tab:soft-labels-table-appx}
\end{table}
\section{Datasets}\label{datasets-appendix}
Here we lay out additional information about the curated datasets, how they were created and their properties. To comply with all methods, we calculate accuracy on the train portion of the data.

\begin{itemize}
\item \textbf{CondInd}: A dataset conforming to the structure of the conditional independence model's classifiers, for assessing the ability to recover its parameters by the iRBM. For the first four classifiers, the probability of predicting the correct label $k$ given $k$ is the true label ($Pr(X_i=k|Y=k)$, the diagonal of the confusion matrix) is sampled uniformly from the interval $[\frac{K-1}{K}, 1]$, where $K=3$ is the number of classes. This ensures they are significantly better than random guessing ($1/K$). The remaining probability mass is distributed uniformly among the incorrect classes. The remaining classifiers in the ensemble are pure noise, with a fixed probability of $1/K$ for all classes.

\item \textbf{Tree3K}: The label Y was sampled from a Uniform(1/k) distribution for $k=3$; each node in the intermediate and bottom layer was generated from his parent with a probability $\psi$ sampled uniformly from [0.7,1] to have the same label as its parent and the remainder sum 1-$\psi$ were uniformly sampled and divided for the rest of the classes, resulting in a $KxK$ probability matrix for each node with its $\psi$ as the diagonal. The tree structure is of dimensions [1,3,4], starting at the true label Y, resulting in $d=12$ classifiers.
\item \textbf{MNIST-E}: The MNIST\citep{mnist-paper} train data of $n=60,000$ samples was used, from which a train-test split of [0.3,0.7] was taken to train 3 classifiers, available from scikit-learn package: A decision tree of depth 5, Gaussian Naive Bayes and a multi-layer Perceptron with $\alpha=1$ and a maximum iteration number of 100. Every classifier predict on the full $n$ samples, then each of which was duplicated 5 times and then 10 percent of its predictions were uniformly randomized, to create a 5-3-1 dependence structure between classifiers, $d=15$.
\item \textbf{PetFinder, CS:GO}: These datasets provides benchmark for data that spans across different modalities, with both tabular, textual and visual fields. CSGO and HS original data is from \cite{lu2023mug}, and PetFinder is from \cite{petfinder-adoption-prediction}. We trained different uni and duo-modal predictors using the AutoGluon~\citep{erickson2020autogluon} library, for a total time of one hour per dataset, to create a diverse set of learners algorithms, namely a text Transformer, tabular MLP, CNN, and purely tabular methods like XGBoost, CatBoost and LightGBM. We used the default train/test split for training the classifiers. 
\item \textbf{UCI Datasets}: The rest of the datasets, are retrieved from the TabRepo~\citep{salinas2023tabrepo} dataset. Originally made for tabular model evaluation on UCI~\citep{dua2019uci} datasets, this benchmark consists of 1310 models from 10 different families, each trained with different hyperparameters and configurations. In order to create a varied ensemble, we choose one classifier from each family at random, and concatenated the predictions to form our datasets ensembles. We then took the five biggest datasets (in terms of examples) for our evaluation. 

\end{itemize}

Wherever a train/val/split was unavailable, test data for both datasets was taken as the last 20 percent of the full data, and 10 percent of the remaining samples were left out as validation.

We used and modified all following datasets in this work. The PetFinder dataset is provided under the CC BY 4.0 license. The CSGO dataset from the MuG benchmark is released under the CC BY-NC-SA 4.0 license. The UCI Machine Learning Repository datasets are available under CC BY 4.0, unless otherwise stated per individual dataset. We used classifiers data released from TabRepo github, following all terms of use as provided. Our amended versions of these datasets are released under a CC BY 4.0 license to ensure free use with proper attribution. Full documentation and dataset versions are provided our code repository and online.

\subsection{ImageNet}\label{imagenet}
To evaluate the scalability of DEEM on large-scale datasets, we constructed a special ImageNet~\citep{imagenet} ensemble dataset. This setting is particularly challenging, as it involves two orders of magnitude more classes than any dataset considered in prior work. Specifically, the dataset comprises $K=1000$ ImageNet classes (compared to a maximum of 30–50 classes in prior benchmarks). We curated this dataset by aggregating predictions from a collection of five pretrained models on the ImageNet validation set. The pretrained models were taken from PyTorch vision package, spanning different architectures and top-1 accuracies between \%82-88, approximately.

To increase its difficulty and relevance, we further filtered out examples where all models unanimously agreed, ensuring that the remaining samples probe how methods leverage hidden knowledge in ambiguous or non-trivial cases. The filtered dataset has 20 percent of the original examples, where the weakest and strongest modified learners have now accuracies of \%38.97 and \%60.88, respectively. Other preprocessing steps were conducted in similar fashion to the unsupervised ensemble datasets above. Unfortunately we were not able to run the official EBCC implementation on this dataset.

\section{MoE Analysis} \label{analysis-appendix}
\begin{table*}[!b]
    \centering
    \small
    \caption{Separate expert subset and remaining data accuracy.}
    \label{table:moe-draft}
    \resizebox{\textwidth}{!}{%
    \small
    \setlength{\tabcolsep}{4pt} 
    \renewcommand{\arraystretch}{0.9} 
    \begin{tabular}{lccccllcc}
    \toprule
    & \multicolumn{2}{c}{MnistE-47} & \multicolumn{2}{c}{MnistE-568} &  \multicolumn{2}{c}{MnistE-4/7}& \multicolumn{2}{c}{AmpData} \\
    Method & Expert& Remaining& Expert& Remaining& Expert&Remaining& Expert& Remaining\\
    \midrule
    MV    & 91.66 & 83.02 & 91.66 & 83.02  & 84.82 &82.98 
& 54.82 & \underline{99.63} \\
 L-SML& 94.87& 86.74& 95.83& 87.55 & 88.32 &86.20 
& 46.34&\textbf{99.92}\\
    DS    & \textbf{97.3} & 92.01 & 95.07 & \underline{93.62}  & 91.70 &\underline{92.52}
& \underline{89.29} & 97.64 \\
    LA$_{1pass}$    & 94.33\,\unc{0.05} & 86.55\,\unc{0.1} & 95.02\,\unc{0.01} & 88.55\,\unc{0.07}  & 87.96 \unc{0.03}&86.51 \unc{0.02}& 28.15\unc{0.05}& \underline{99.91}\unc{0.14}\\
    LA$_{2pass}$   & 95.09\,\unc{0.0} & 86.80\,\unc{0.18} & 95.37\,\unc{0.0} & 89.57\,\unc{0.01}  & 88.38 \unc{0.0}&86.73\unc{0.0}& 28.15\unc{0.05}& \underline{99.91}\unc{0.11}\\
    EBCC  & 96.68 & 88.60 & \underline{95.74} & 90.77  & \underline{94.53}&88.57 
& 34.06 & 98.70 \\
    HLM   & 92.82 & 85.38 & 92.59 & 87.19  & 87.11 &85.28 
& 60.28 & 98.36 \\
    \midrule
 iRBM & 96.62 \unc{0.07} & 86.32 \unc{0.18}& 96.07 \unc{0.10}& 88.89\unc{0.09} & 92.28 \unc{0.39}&86.13\unc{0.25}& 29.99\unc{0.98} &\textbf{99.92}\unc{0.01}\\
    \textbf{DEEM} & \underline{97.1}\,\unc{0.5} & \textbf{94.04}\,\unc{0.61} & \textbf{96.07}\,\unc{0.42} & \textbf{94.29}\,\unc{0.76}  & \textbf{95.27}\unc{0.08}&\textbf{94.69} \unc{0.0}& \textbf{96.63}\,\unc{0.44} & 96.17\,\unc{0.81} \\
    \bottomrule
    \end{tabular}
    }
\end{table*}
\begin{figure}
    \centering
    \includegraphics[width=0.65\textwidth]{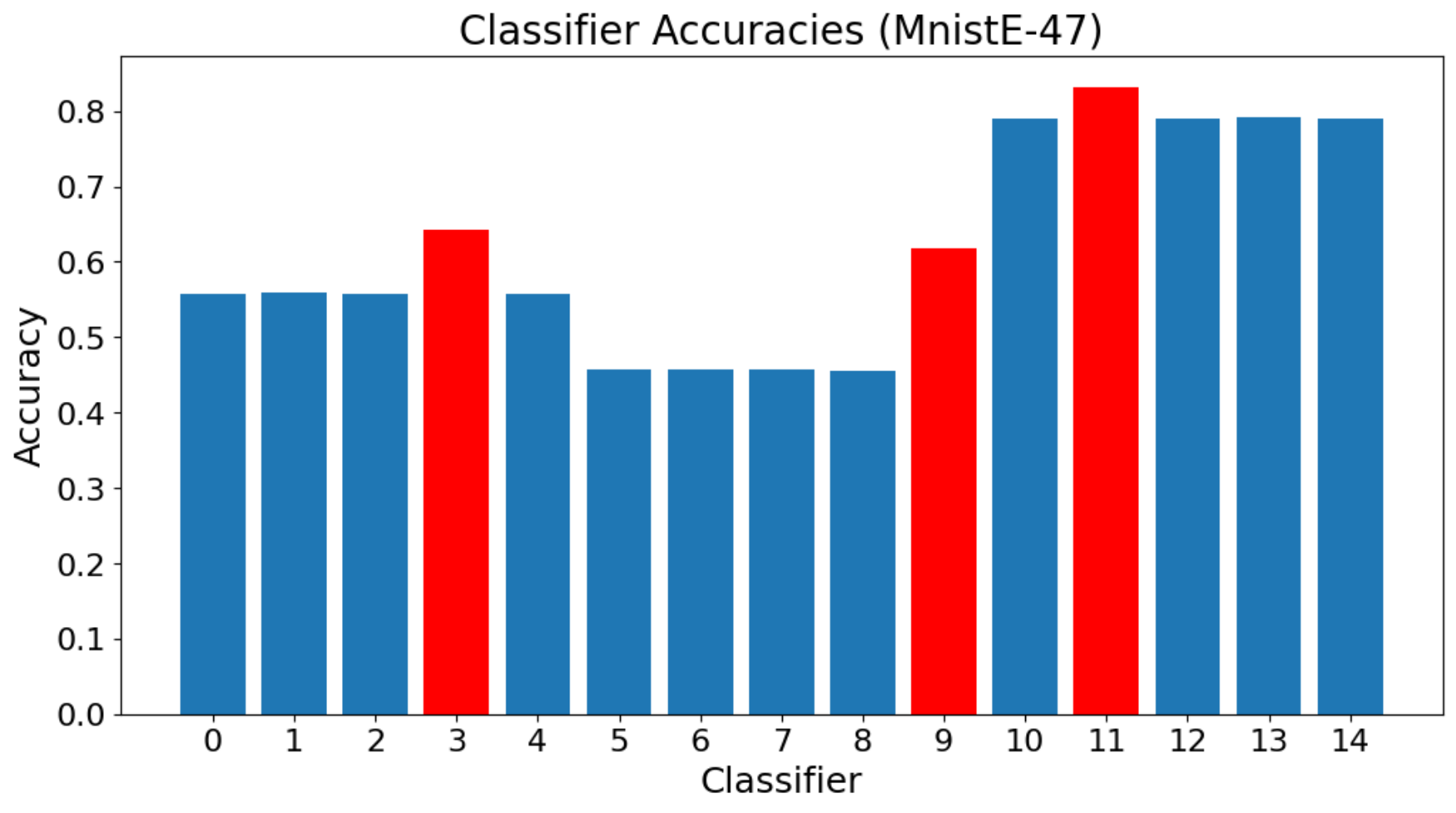}
    \caption{MnistE-47 classifiers' individual accuracies on the full dataset.}
    \label{fig:mniste47-accs}
\end{figure}
\begin{figure}
    \centering
    \includegraphics[width=0.65\linewidth]{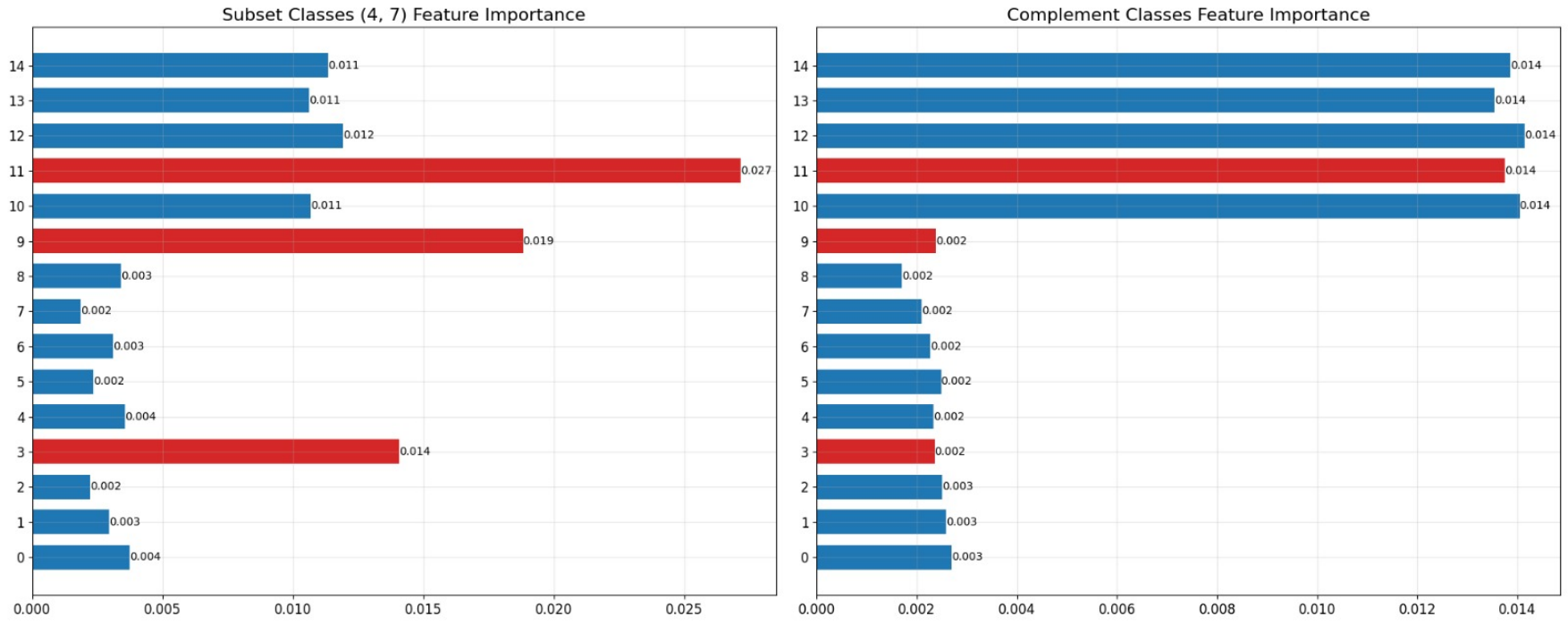}
    \caption{MnistE-568 classifiers' individual accuracies on the full dataset.}
    \label{fig:mniste568-accs}
\end{figure}
Here we present further details of our MoE experiement for the DEEM model. For the MoE datasets we have in total one simulated dataset (AmpData) and three real-world datasets (MnistE47, MnistE568, MnistE4/7):

\begin{itemize}
    \item AmpData: Our dataset consists of $d=6$ classifers, and $K=5$ classes. The 5 first classifiers were assigned \%90-92 accuracy on the last three classes, and score random on the first two classes. The last classifier amplifies the shared information in the ensemble, by having oracle performance on the first two classes, while random guessing on the last three classes. This scenario is designed to test added specialization when knowledge overlap is minimal.
    \item MnistE-47 and 568 Datasets: We altered the MnistE dataset in the following way: To reduce classifiers' dependence, we first introduced some label noise, selecting 10\% of examples at random and assigning random labels to them. Then, we selected 3 classifiers, one from each classifier 'family', and gave it oracle performance on given classes, 4 and 7 for the first dataset, and 5,6 and 8 for the second dataset. These tests allows to understand if a model is able to utilize the specialization even when knowledge overlap between the experts and rest is less pronounced. These tests are similar but with different oracle classes.
    \item MnistE-4/7: This dataset was curated by using the MnistE dataset in similar way to the above datasets, but here only two classifiers were selected, and each classifier was given oracle performance on only one distinct class, classes 4 and 7 respectively. This setting was designed to assess the ability to utilize multiple distinct sources of specialization. More detailed information is available in subsection ~\ref{distinct-moe} below.
\end{itemize}

Every trial was run 5 times with the same steps procedure as other datasets in this paper. Figures~\ref{fig:mniste47-accs} and~\ref{fig:mniste568-accs} show the accuracies of every individual classifier of the altered MnistE datasets, with the experts marked in red.

\begin{figure}
    \centering
    \includegraphics[width=0.75\linewidth]{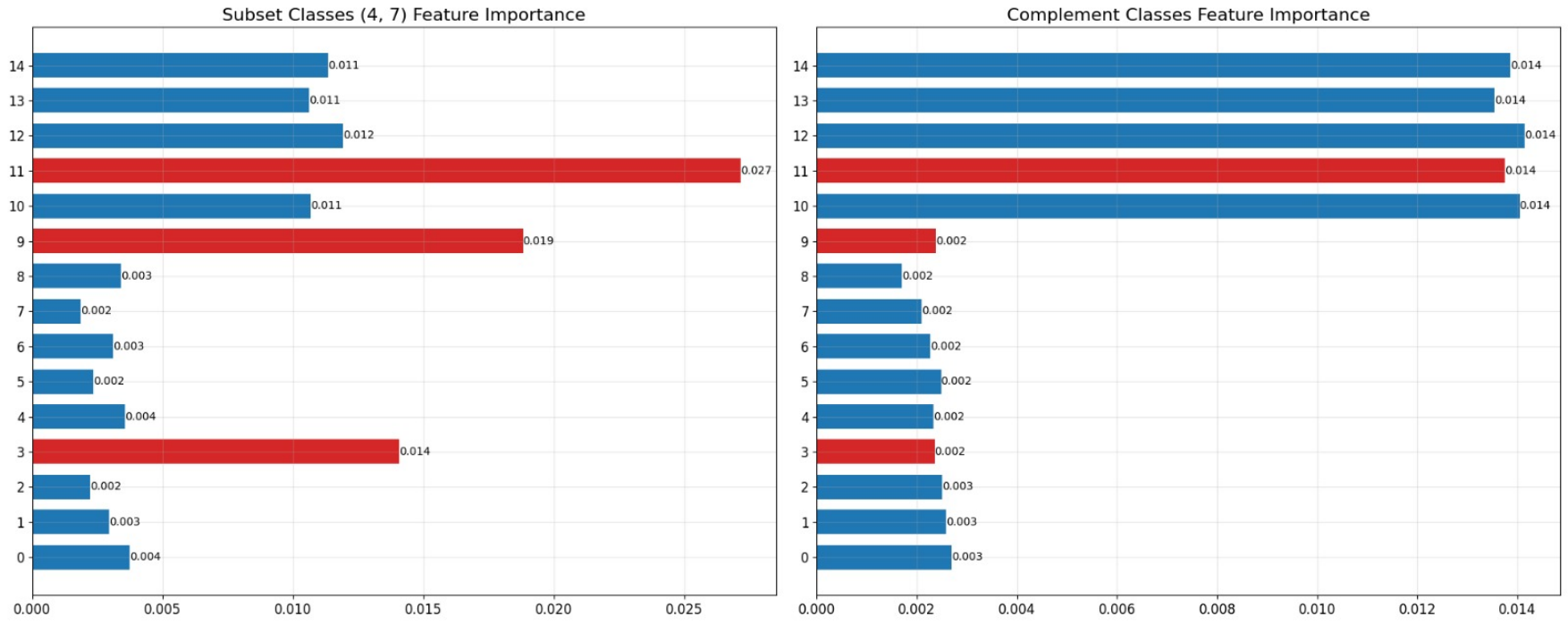}
    \caption{MnistE-47 leaners impact. The left panel shows impact on the subset, while the right panel shows for the rest of the data.}
    \label{fig:mniste47-impact}
\end{figure}

The learners impact on DEEM prediction of the MnistE-47 dataset is shown in Figure~\ref{fig:mniste47-impact}, and as seen in Figure~\ref{fig:moe}, DEEM was able to increase the specialized learners effect on the expert portion of the data without hurting their contribution on the remaining examples.

\subsection{Two distinct areas of expertise}\label{distinct-moe}
In the additional setup of MnistE-4/7, we altered the MnistE dataset to construct a scenario, where only two experts are chosen instead of three, but one has oracle access only to the 4 class, and the second with an access to the 7 class predictions only. This further assess fusion of knowledge from multiple distinct sources of expertise. The results are presented below:
\begin{table}[h]
\caption{Performance on MnistE-4/7 dataset.}
\centering
\begin{tabular}{lcccc}
\toprule
\textbf{MnistE-4/7} & \textbf{Expert}& \textbf{Remaining}& \textbf{Class 4 Alone} & \textbf{Class 7 Alone} \\
\midrule
MV      & 84.82 & 82.98 & 81.08 & 88.33 \\
 L-SML   & 88.32 & 86.20 & 86.48 &90.04 \\
DS      & 91.70 & \underline{92.52}& 85.72 & \textbf{96.83} \\
LA$_1$pass & 87.96 $\pm$ 0.03 & 86.51 $\pm$ 0.02 & 85.19 $\pm$ 0.02 & 90.55 $\pm$ 0.04 \\
LA$_2$pass & 88.38 $\pm$ 0.0 & 86.73 $\pm$ 0.0 & 85.90 $\pm$ 0.0 & 90.71 $\pm$ 0.01 \\
EBCC    & \underline{94.53}& 88.57 & \underline{92.72}& \underline{96.23}\\
HLM     & 87.11 & 85.28 & 85.09 & 89.00 \\
\midrule
iRBM    & 92.28 $\pm$ 0.39 & 86.13 $\pm$ 0.25 & 90.42 $\pm$ 0.56 & 94.03 $\pm$ 0.23 \\
DEEM    & \textbf{95.27 $\pm$ 0.08} & \textbf{94.69 $\pm$ 0.0} & \textbf{94.64 $\pm$ 0.0} & 96.15 $\pm$ 0.0\\
\bottomrule
\end{tabular}
\label{tab:mniste}
\end{table}
We have additionally tested the performance on each of the classes separately (Class 4/7 Alone), to better understand the impact of utilizing each expertise on its own. These results suggest DEEM is able to successfully fuse distinct expertise without overriding one atop of the other, while maintaining high performance across the rest of the dataset. 

\section{Classifier Scalability}
We have conducted an additional experiment to test classifier scalability. We expanded the GesturePhsm dataset in one order of magnitude, from 11 to 100 classifiers (generating 8 variations of each distinct classifier with independent random noise, and completing the remainder with random-guess classifiers). 

DEEM successfully converged and maintained superior accuracy over the baseline (DS), demonstrating that the method scales robustly with the classifier size.

\begin{table}[!h]
    \centering
    \caption{Accuracy over extended GesturePhsm100 dataset, with 100 classifiers.}
    \begin{tabular}{cccccccccc}\toprule
         BestClf & AvgClf   & MV  & L-SML & DS    & EBCC  & HLM   & LA$_{2pass}$ & iRBM   & DEEM\\\midrule
         69.8    & 52.55 $\pm$ 13.27 & 63.89 & 65.06 & 65.12 & \underline{66.17} & 66.12 & 64.08 & 64.09 $\pm$ 0.02 & \textbf{67.05} $\pm$ 0.12 \\ \bottomrule
    \end{tabular}
    
    \label{tab:classifier-scalability}
\end{table}
\section{Sensitivity Assessment}\label{sensitivity-appendix}
Here we present a small preliminary sensitivity assessment for the DEEM model.

Our goal in this preliminary assessment is to elucidate how classifiers' quality might impact DEEM's prediction. In other words, how the model responds to a change in the ensemble, reacting to the addition of classifiers that benefit the prediction and those who are pure noise. This Classifier Impact assessment results are averaged over 3 runs.

Here, we start by taking the MnistE dataset, constructing a new $d=3$ base ensemble, made from three classifiers, one from each algorithm family. Then, for the quality classifiers, we simply add different classifiers from the original dataset, where for the non-quality ones, we construct random-guess classifiers with a small $\varepsilon=0.001$ percent of examples assigned the true labels, so the non-quality classifiers are better than random (following the conditions presented in Remark \autoref{remark:iden}. ). The good classifiers are added in alternating order, where every addition is from a subsequent algorithm family.

We note that the three families of classifiers have different performance, where the last family achieves around 20\% increase in accuracy over the other two families, which might affect the overall performance of the ensemble differently.
\begin{figure}[b!]
    \centering
    \includegraphics[width=0.6\linewidth]{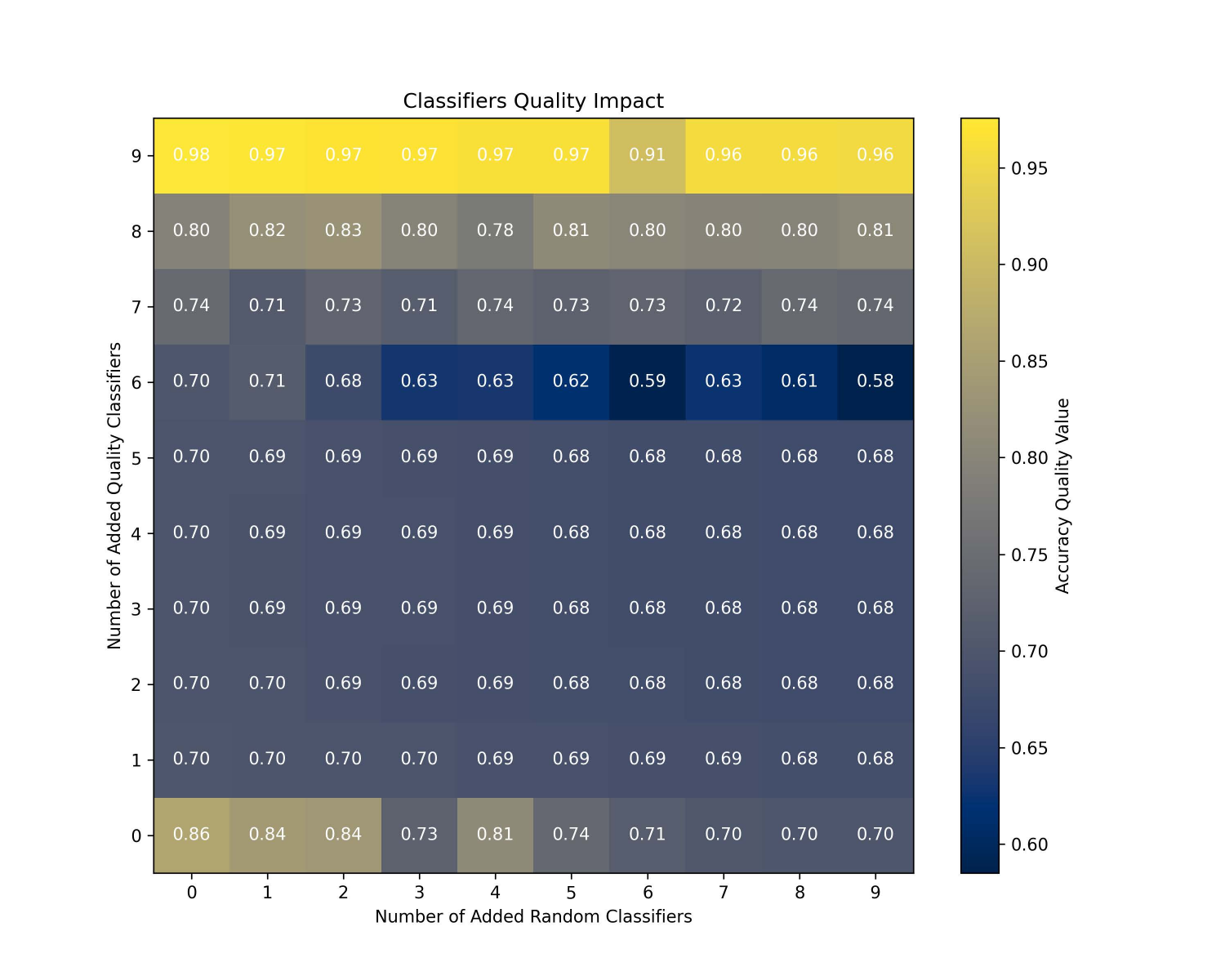}
    \caption{Impact of added learners on the model's accuracy quality. The axes show the amount of added high and low quality learners to the original ensemble.}
    \label{fig:quality-impact}
\end{figure}

Since the accuracy threshold of the individual classifiers and the majority vote changes from ensemble to ensemble, we employ a slightly different metric we call 'accuracy quality': We mark examples where the true label is among one of the example's prediction, and after the model is trained we see how many of these examples the model got right, yielding a subset of all data. This allows to have a more standardized metric for our assessment, examining the extent to which DEEM was able to decipher the true answer.

As depicted in \autoref{fig:quality-impact}, When non-quality classifiers were added, the model keeps its accuracy quality and handles the noise robustly, with slight decreases in the 0 and 6th rows, which we believe is due to instability of the training as mentioned before, given we ran all ensembles with the same hyperparameters. When we add good classifiers, performance remains the same as the overall averaged accuracy remains the same, yet a steady increase in accuracy quality is shown, with 4\%, 7\% and 10\% increase respectively for each subsequent row, starting from the 7th row.

Combining with the previous results, we hypothesize that in order to model the complex inter-relationships between learners in the most potent way, a certain level of persistent true signal must exist within the ensemble for DEEM to perform the most effective. However, the exact amount of this persistent signal remains unknown and is subject to further research.
\section{Disentanglement Through the Multinomial Layers}\label{disentaglement-multinomial-draft}
Following the addition of the multinomial layers to assist with the reduce of the conditional dependencies between classifiers, we empirically validate the claim using the following analysis: 

At each layer of DEEM, and for each true label (Importantly, the true label is used only for analysis purposes, not for training), we computed the mutual information (MI) between classifier outputs as a proxy for conditional independence. The input to the MI computation is the argmax prediction of each classifier (feature), treated as a discrete random variable.\footnote{We compute the MI using mutual\_info\_score from scikit-learn, which performs binning via empirical co-occurrence counts for compting the value.}

These analyses were conducted on a trained DEEM with two layers on the MnistE dataset. For each true class and each layer, we construct a full $d \times d$ MI matrix which we summarize below using the maximum MI and Frobenius norm (Excluding diagonal entries).

\begin{figure}[h]
    \centering
    \includegraphics[width=0.5\linewidth]{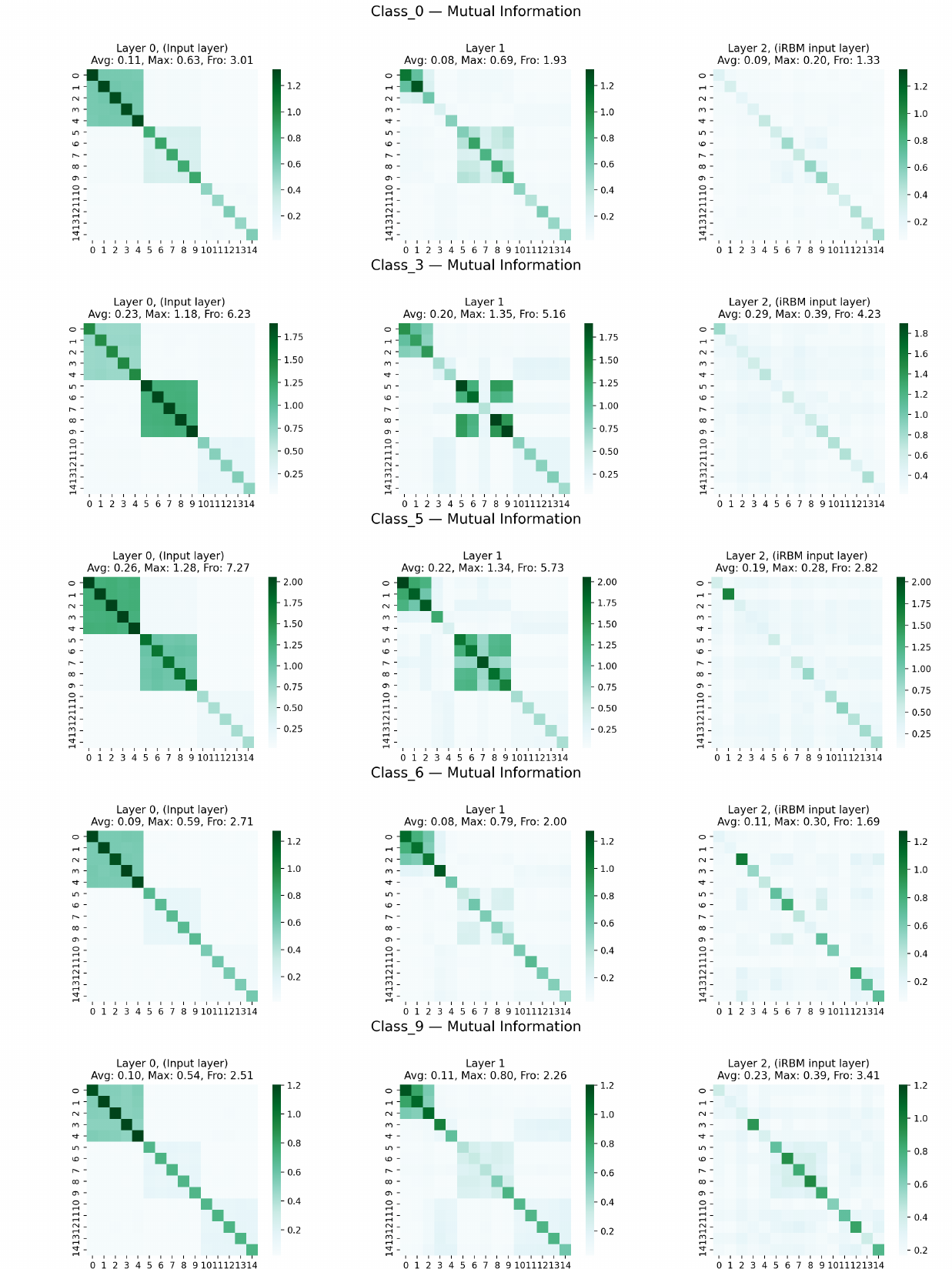}
    \caption{The mutual information matrices of a trained DEEM with 2 multinomial layers. Each row corresponds to a specific true label class subset of the data. Starting from the input layer in the leftmost column, it can be seen how the mutual information is gradually being reduced as we progress through the multinomial network, until we get disentangled features as input for the iRBM component.}
    \label{fig:maxmi_frob_apx}
\end{figure}

\begin{table}[h]
\centering
\begin{tabular}{lcc}
\toprule
\textbf{Layer} & \textbf{Max MI} & \textbf{Frobenius Norm} \\
\midrule
Layer 0 (Input)      & $0.827 \pm 0.294$ & $4.428 \pm 1.796$ \\
Layer 1              & $0.981 \pm 0.272$ & $3.504 \pm 1.439$ \\
Layer 2 (iRBM input) & $0.314 \pm 0.068$ & $2.687 \pm 0.985$ \\
\bottomrule
\end{tabular}
\caption{Max MI and Frobenius Norm across layers.}
\label{tab:maxmi_frob-draft}
\end{table}
These statistics reflect an approximation for the level of dependence among classifiers at each layer. As shown in Table~\ref{tab:maxmi_frob-draft} and Figure~\ref{fig:maxmi_frob_apx} , both max MI and Frobenius norm decrease across layers, suggesting that the learned representations are shifted towards conditional independence as we progress through the deep multinomial network. 
These results support our hypothesis, which is as data flows through the network, classifier outputs become less statistically dependent. The drop in MI is particularly pronounced in the final layer where the iRBM start to operate, suggesting that the network learns to shape representations to help overcome the iRBM structural assumption of independence. 

\section{Considerations}\label{considerations-appendix}
When considering limitations for this work, we mostly focus on the possibility to acquire the true posterior of the iRBM and the remarks we outline in Section \ref{subsec:irbm}, and the traditional difficulty in training energy-based models in general \citep{song2021train}. Regarding the boundaries of our proof, the ensemble conditions suggested by \citep{chang1996full} are realistic in most real-world scenarios, but knowing how much data is sufficient to acquire the true posterior accurately enough remains a question. 

However, with the growing abundance of available data, this is relaxed to a degree. As for training the iRBM model, as have been shown in the experiments, we are able to train it successfully and consistently, given a proper initialization and a gradient sampling method.

Other related issues stem from the aforementioned instability of energy-based models training. This can create some scenarios where the model diverges without noticeable indications. Although it is not guaranteed to converge to the most optimal result, our initialization scheme and suggestions regarding hyperparameters can easily assess when a model has collapsed, and choose a competitive LR that is both learn and stable.

In future work, several avenues remains open for exploration. First, other mechanisms or deep architectures can be applied to relax the conditional independence assumption, as well as better training control and optimization, in order to improve its effectiveness and robustness.  

\paragraph{Scalability} Most works in this field, including this paper (and arguably most real world applications of ensemble learning problems) consider only medium sized multi-class problems, both regarding dataset size and number of classes. Since our method trains by mini-batches, it allows easy processing of large scale datasets. However, due to the one-hot nature of the inputs of RBMs, DEEM deep layers and iRBM weights are both dependent on the number of classes, and grow quadratically as they increase. High number of classes might incur heavy memory overhead for intermediate matrices multiplication, specifically in computing the categorical distribution of the DLP sampler in Eq.~\ref{eq:dlp-q}, due to its intermediate proposal matrix that is sized $d\cdot K^3$. This might amplify the inherent EBM training instability and sparsity input data concerns. In that case, memory considerations need to be taken into account, which is subject to further research. For a temporary mitigation, one can compute the matrices multiplication in chunks and resort to smaller mini-batches or use gradient accumulation if possible. 

\paragraph{Ethic statement.} In whole, we believe DEEM allows for positive implementation in many areas. Malicious use can not ever be avoided completely, but as our theme setting assumes the absent of any data or model information, let alone one that can jeopardize one's privacy or security, and learns the classifiers' dependencies in an unsupervised manner, we believe it brings the possibility of unintended use to a minimum.

\clearpage

\end{document}